\documentclass[lettersize,journal]{IEEEtran}
\usepackage{cite}
\usepackage{moreverb,url}
\usepackage{graphicx}
\usepackage{booktabs}
\usepackage{threeparttable}
\usepackage{caption}
\usepackage{siunitx}
\usepackage{multirow}
\usepackage{pifont}
\usepackage{amssymb}
\usepackage{amsmath}
\usepackage{tabularx}
\usepackage{comment}
\usepackage{mathtools} 
\usepackage{stmaryrd}
\usepackage[table,dvipsnames]{xcolor}
\usepackage{arydshln}
\usepackage[normalem]{ulem} 
\newcommand{\ul}[1]{\uline{#1}}
\usepackage{tikz} 
\newcommand\BibTeX{{\rmfamily B\kern-.05em \textsc{i\kern-.025em b}\kern-.08em
T\kern-.1667em\lower.7ex\hbox{E}\kern-.125emX}}
\usepackage{listings}
\lstdefinelanguage{json}{
    basicstyle=\ttfamily\footnotesize,
    numbers=left,
    numberstyle=\tiny,
    stepnumber=1,
    numbersep=5pt,
    showstringspaces=false,
    breaklines=true,
    literate=
     *{0}{{{\color{black}0}}}{1}
      {1}{{{\color{black}1}}}{1}
      {2}{{{\color{black}2}}}{1}
      {3}{{{\color{black}3}}}{1}
      {4}{{{\color{black}4}}}{1}
      {5}{{{\color{black}5}}}{1}
      {6}{{{\color{black}6}}}{1}
      {7}{{{\color{black}7}}}{1}
      {8}{{{\color{black}8}}}{1}
      {9}{{{\color{black}9}}}{1}
      {:}{{{\color{black}{:}}}}{1}
      {,}{{{\color{black}{,}}}}{1}
      {"}{{{\color{red}{"}}}}{1},
}
\usepackage[colorlinks,bookmarksopen,bookmarksnumbered,citecolor=green,urlcolor=black]{hyperref}

\definecolor{C1}{HTML}{ECA8A9}
\definecolor{C2}{HTML}{74AED4}
\definecolor{C3}{HTML}{D3E2B7}
\definecolor{C4}{HTML}{CFAFD4}
\definecolor{C5}{HTML}{F7C97E}
\definecolor{C6}{HTML}{E8D5C4}
\definecolor{C7}{HTML}{EEEEEE}
\definecolor{C8}{HTML}{BCEE68}

\newcommand{\cmark}{\ding{51}}%
\newcommand{\xmark}{\ding{55}}%

\definecolor{tabfirst}{rgb}{1, 0.75, 0.7}
\definecolor{tabsecond}{rgb}{1, 0.85, 0.65}
\definecolor{tabthird}{rgb}{1, 0.96, 0.7}
\definecolor{cvprblue}{rgb}{0.21,0.49,0.74}

\begin{document}

\title{GaussianDWM++: Language-Grounded 3D Gaussian Driving World Model for Unified Scene Understanding, Editing, and Multi-Modal Generation}
\author{Tianchen Deng, Xuefeng Chen, Shuang Wu, Qu Chen, Jiajun Zhu, Bo Dai,\\ Jianfei Yang, Hesheng Wang,~\IEEEmembership{Senior Member,~IEEE}}
\maketitle

\makeatletter
\begingroup
\renewcommand*{\@makefnmark}{}
\footnotetext{Tianchen Deng and Hesheng Wang are with the School of Automation and Intelligent Sensing, Shanghai Jiao Tong University; the State Key Laboratory of Avionics Integration and Aviation System-of-Systems Synthesis; and the Shanghai Key Laboratory of Navigation and Location Based Services, Shanghai 200240, China. Xuefeng Chen and Shuang Wu are with Tsinghua University, Shenzhen, China. Qu Chen, Jiajun Zhu, and Bo Dai are with Chongqing Afari Intelligent Drive Co., Ltd., China. Jianfei Yang is with Nanyang Technological University, Singapore. This work was supported by the National Key R\&D Program of China (Grant No.~2024YFB4708900) and in part by the National Natural Science Foundation of China under Grants 62225309, U24A20278, and 62361166632. The first three authors contributed equally to this work. Corresponding author: Hesheng Wang (e-mail: wanghesheng@sjtu.edu.cn).}
\endgroup
\makeatother

\markboth{IEEE Transactions on Pattern Analysis and Machine Intelligence}%
{Deng \MakeLowercase{\textit{et al.}}: GaussianDWM++}

\IEEEpubidadjcol

\begin{abstract}
Driving World Models (DWMs) have recently advanced rapidly with generative models, yet most existing methods mainly focus on conditional scene generation and lack explicit 3D scene understanding, language-grounded reasoning, and controllable 4D editing capabilities. Moreover, commonly used point cloud, occupancy, or BEV representations make it difficult to achieve fine-grained alignment between textual information and the underlying 3D scene structure. To address these limitations, we propose a foundation-feature Gaussian driving world model that unifies scene understanding, language-grounded reasoning, controllable 4D editing, and multi-modal generation within a single framework. Specifically, we introduce a foundation-feature Gaussian tokenizer that directly distills Qwen visual-language features into 3D Gaussian primitives, building a compact open-vocabulary Gaussian semantic field. We further design a geometry-aware Gaussian adapter that combines importance-aware hierarchical selection with text-conditioned Perceiver-style cross-attention to aggregate dense Gaussian primitives into compact world tokens. To improve representation compatibility, we introduce a KL-based Gaussian--image distribution alignment objective that aligns Gaussian world tokens with foundation image tokens. Based on the aligned Gaussian representation, our framework further supports instruction-controllable scene editing, including weather-conditioned generation and dynamic vehicle manipulation, while preserving temporal, geometric, and semantic consistency. We further construct NuScenes-GSQA, a large-scale dataset containing 1000 nuScenes driving scenes represented by foundation-feature 3D Gaussians, together with Gaussian-aware QA annotations derived from NuInteract. Extensive experiments on broader driving benchmarks demonstrate that our method achieves state-of-the-art performance across scene understanding, visual grounding, planning-oriented reasoning, and controllable 4D generation tasks. We will release the code and datasets publicly on \href{https://github.com/dtc111111/GaussianDWM}{https://github.com/dtc111111/GaussianDWM}.
\end{abstract}

\begin{figure}[t]
    \centering
    \includegraphics[width=\linewidth]{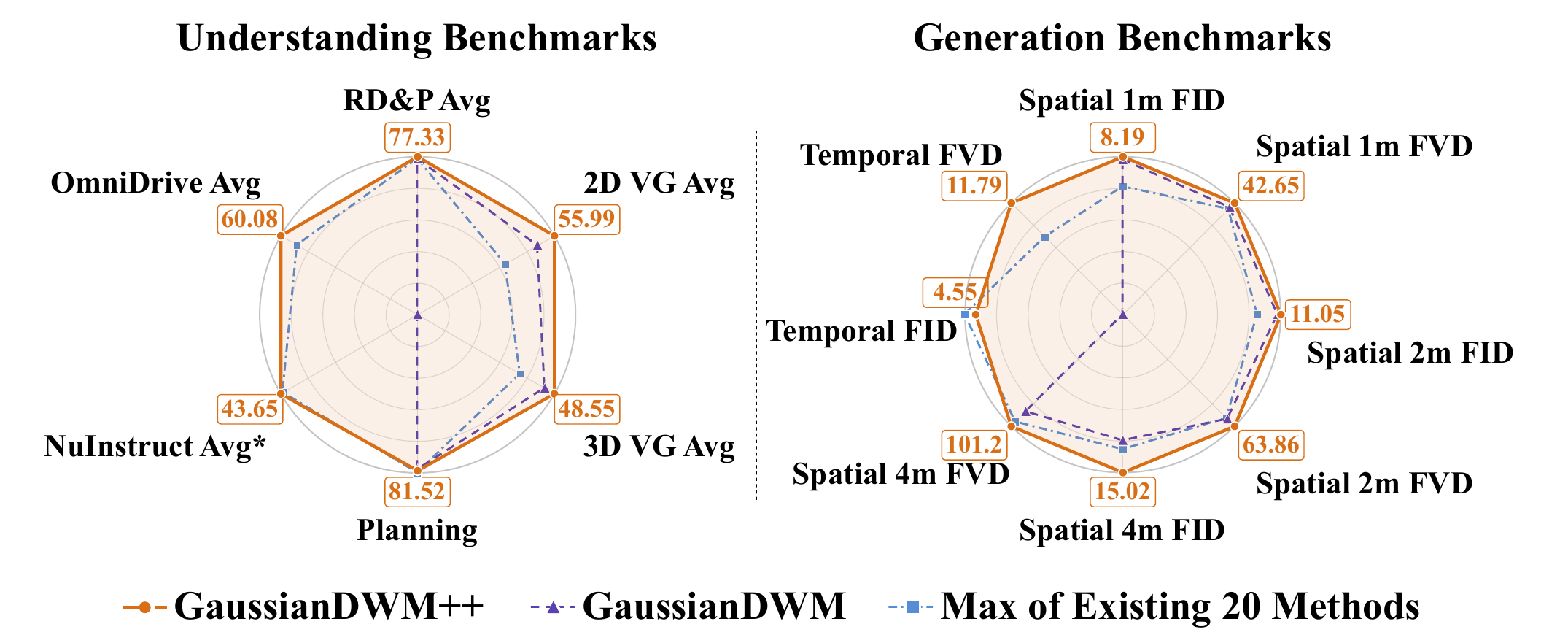}
    \caption{\textbf{Overall performance on understanding and generation benchmarks.}
GaussianDWM++ achieves state-of-the-art performance in both scene understanding
and scene generation. Radar radii are normalized as
$\mathrm{Metric}_i/\mathrm{Metric}_{\max}$ for understanding and
$\mathrm{Metric}_{\min}/\mathrm{Metric}_i$ for generation, so that larger radii
indicate better performance. Boxed values show the original GaussianDWM++ results.}
    \label{fig:radar}
\end{figure}

\begin{figure*}[t]
    \centering
    \includegraphics[width=\linewidth]{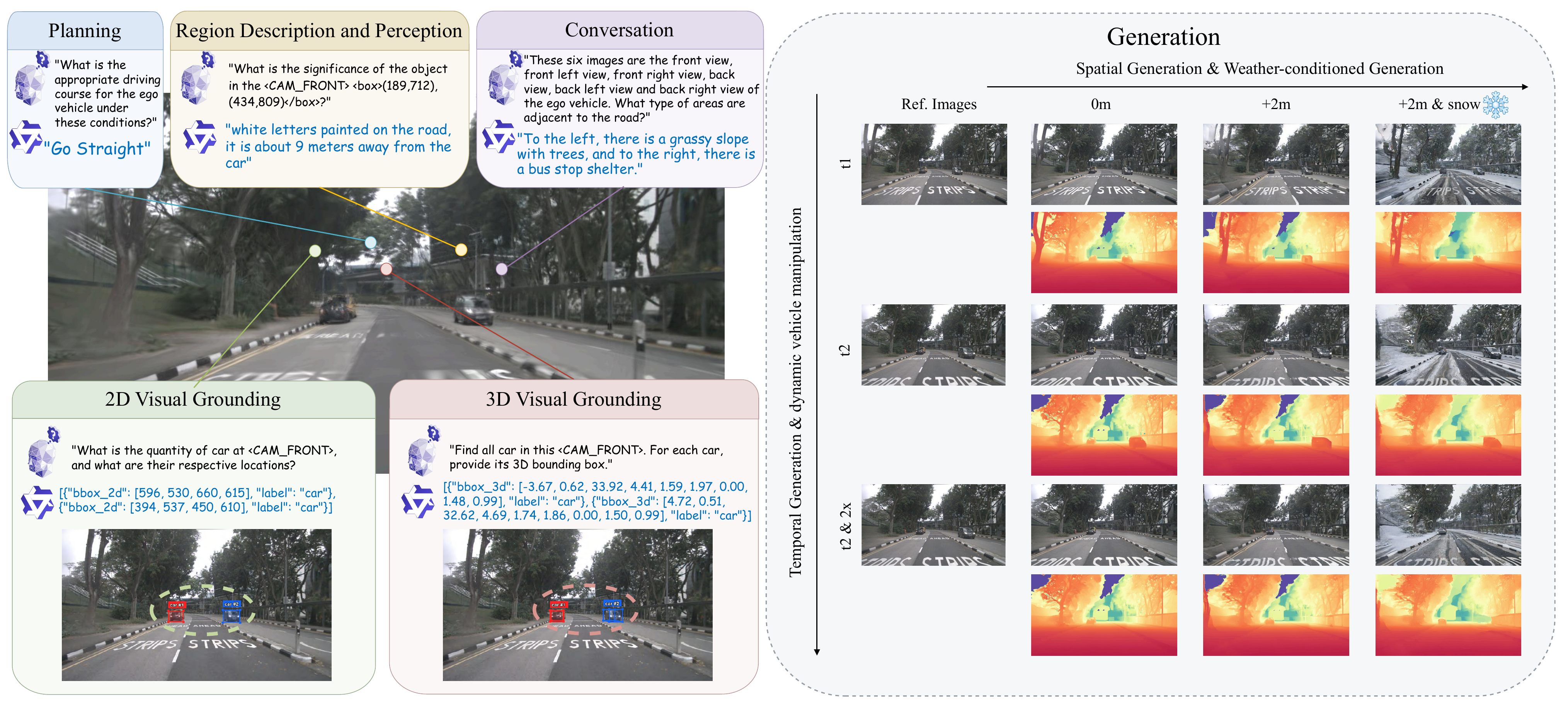}
    \caption{We propose the unified 3D Gaussian-based world model framework
that achieves comprehensive scene understanding and scene generation
for driving scenarios.
It efficiently encodes complex scenes, samples task-relevant information,
and handles diverse question-answering tasks.
Moreover, by leveraging the extracted world knowledge,
our framework guides the generative model to perform accurate
spatial and temporal scene generation.}
    \label{fig:teaser}
\end{figure*}

\section{Introduction}

Driving World Models (DWMs)~\cite{vista,gaia,drivingwm} have become increasingly important for autonomous driving, owing to their ability to predict future scene evolution, synthesize realistic driving scenarios, and provide simulation data for risk forecasting, route optimization, and corner-case training. Existing DWMs mainly focus on generating future observations conditioned on input data, such as images~\cite{drivedreamer}, videos, point clouds~\cite{pointcloud}, or BEV/occupancy representations~\cite{uniscene}. These modalities effectively describe the visual or geometric evolution of driving environments, but they are still limited in their ability to interpret, describe, and reason about the underlying scene. As a result, most existing DWMs remain difficult to query with natural language and cannot directly provide contextual understanding for tasks such as visual question answering, 2D/3D grounding, planning-oriented reasoning, or instruction-conditioned scene editing.

Recent advances in Large Language Models (LLMs) and Vision-Language Models (VLMs)~\cite{qwen3} have demonstrated strong capabilities in semantic understanding, commonsense reasoning, and language-grounded visual perception. These developments suggest a promising direction for autonomous driving: combining the generative capability of DWMs with the scene understanding and reasoning capability of VLMs. Pioneering works such as HERMES~\cite{hermes} have explored this direction by unifying scene understanding and scene generation within driving world models. However, these methods typically rely on BEV, depth, or occupancy features to represent 3D spatial information. Such representations are effective for structured driving perception, but they often establish only feature-level alignment between language and space, making it difficult to achieve fine-grained correspondence between textual concepts, image observations, and explicit 3D scene geometry.

To address these limitations, we propose \textbf{GaussianDWM++}, a novel foundation-feature 3D Gaussian driving world model that unifies scene understanding, language-grounded reasoning, controllable 4D scene editing, and multi-modal generation within a single 3D Gaussian-based framework, as summarized in Fig.~\ref{fig:teaser}. Different from point-, BEV-, or occupancy-based representations, 3D Gaussian Splatting (3DGS) provides an explicit, continuous, and differentiable scene representation, where each Gaussian primitive carries spatial position, anisotropic geometry, opacity, appearance, and semantic attributes. This primitive-level representation offers several advantages for driving world modeling. First, it naturally preserves explicit 3D spatial structure and multi-view consistency. Second, it supports differentiable rendering, enabling image-, depth-, and feature-level supervision from multi-view observations. Third, it provides a flexible carrier for foundation visual-language features, allowing language semantics to be directly embedded into 3D scene primitives rather than being indirectly aligned through BEV or image-level features.

Based on this representation, we first introduce a foundation-feature Gaussian tokenizer that directly distills Qwen visual-language features into 3D Gaussian primitives. Compared with the previous LangSplat-dependent language Gaussian construction, our tokenizer learns a compact and trainable open-vocabulary Gaussian semantic field from multi-view observations. This design directly connects foundation visual-language features with explicit 3D Gaussian primitives, thereby improving the alignment among image observations, language semantics, and 3D spatial geometry.

However, directly feeding all Gaussian primitives into an LLM or a generative model is impractical, since a driving scene may contain tens or hundreds of thousands of Gaussians. Moreover, dense Gaussian primitives are highly redundant, and not all primitives are equally relevant to a given query or generation condition. To overcome this challenge, we design a geometry-aware Gaussian adapter to transform dense Gaussian primitives into compact world tokens. Specifically, we first introduce an importance-aware hierarchical Gaussian selection strategy that evaluates Gaussian primitives according to their geometric quality and driving relevance. We then employ a Perceiver-style cross-attention module in which the pooled text embedding modulates a small set of learned queries. These text-conditioned queries attend to the selected Gaussian tokens and aggregate them into compact LLM-compatible world tokens. This design enables efficient token extraction while preserving task-relevant 3D spatial and semantic information.

To further improve representation compatibility, we introduce a \textbf{KL-based Gaussian--image distribution alignment} objective. Instead of simply projecting Gaussian tokens with independent MLPs, we regularize the compact Gaussian world tokens toward the Qwen image-token distribution from the same scene. This image-guided objective transfers semantically rich foundation features to the Gaussian tokens while preserving their explicit 3D structure. In this way, the proposed adapter not only compresses dense 3DGS primitives into compact tokens, but also makes the resulting tokens more compatible with the visual representation consumed by the LLM.

In addition to the model design, we observe that existing driving language datasets are mainly built upon image-level, BEV-level, or object-level inputs, while large-scale datasets that directly support 3DGS-based scene understanding and reasoning remain scarce. To fill this gap, we construct a large-scale 3D Gaussian driving scene dataset from nuScenes~\cite{nuscenes}, containing 1000 outdoor urban scenes represented with high-quality 3D Gaussian primitives. The dataset provides explicit geometry, appearance, and foundation-feature representations for each scene, enabling generalizable 3D Gaussian-based scene understanding with LLMs. Furthermore, using NuInteract annotations, we build a large-scale 3DGS-based QA dataset covering scene description, visual grounding, spatial reasoning, planning-oriented reasoning, and 3D-aware language interaction. To the best of our knowledge, this is among \textbf{the first large-scale 3D Gaussian datasets} for outdoor urban driving scene understanding with LLMs. This dataset not only supports the training and evaluation of our framework, but also provides a useful benchmark for future 3D Gaussian-based driving world models.

Furthermore, we extend the original dual-condition generation framework to instruction-controllable 4D driving scene editing and generation. The proposed generation module leverages high-level world knowledge from the VLM together with low-level geometric and image conditions from the Gaussian scene representation. Beyond spatial and temporal scene generation, our framework supports more diverse and controllable 4D editing modes, including weather-conditioned generation and dynamic vehicle manipulation. These editing operations are performed while preserving temporal coherence, geometric consistency, and semantic fidelity, enabling the model to function not only as a scene generator, but also as an interpretable and controllable driving world simulator.

Overall, the contributions of our paper are summarized as follows:
\begin{itemize}
\item We propose a novel foundation-feature 3D Gaussian driving world model that unifies scene understanding, language-grounded reasoning, controllable 4D scene editing, and multi-modal generation within a single 3D Gaussian-based framework. To the best of our knowledge, this is the first unified 3D Gaussian driving world model for scene understanding and generation.

\item We introduce a foundation-feature Gaussian tokenizer that directly distills Qwen visual-language features into 3D Gaussian primitives, building a compact and trainable Gaussian semantic field. We design a geometry-aware Gaussian adapter that combines importance-aware hierarchical Gaussian selection with Perceiver-style cross-attention, where pooled text embeddings modulate learned queries to aggregate dense Gaussian primitives into compact world tokens.

\item We propose a KL-based Gaussian--image distribution alignment objective that regularizes Gaussian world tokens toward the Qwen image-token distribution, improving their compatibility with foundation visual representations.

\item We extend the generation module to instruction-controllable 4D driving scene editing and generation, supporting weather-conditioned generation and dynamic vehicle manipulation while preserving temporal, geometric, and semantic consistency.

\item We construct a large-scale 3D Gaussian driving scene dataset, NuScenes-GSQA, containing 1000 outdoor urban scenes from nuScenes represented with high-quality 3D Gaussian primitives. Using NuInteract annotations, we further build a large-scale 3DGS-based question-answering dataset for scene description, visual grounding, spatial reasoning, planning-oriented reasoning, and 3D-aware language interaction. Extensive experiments on broader driving benchmarks demonstrate the effectiveness of our framework on scene understanding and 4D generation tasks. 
\end{itemize}

This paper is an extension of our previous conference version CVPR 2026~\cite{Deng_2026_CVPR}. Compared with the conference paper, this journal version makes several substantial extensions. First, we replace the LangSplat-based language Gaussian construction with a foundation-feature Gaussian tokenizer that directly distills Qwen features into 3D Gaussian primitives. Second, we redesign the Gaussian tokenization module by introducing an importance-aware hierarchical sampling strategy and a Perceiver-style cross-attention adapter, in which pooled text embeddings modulate learned queries for compact and geometry-aware aggregation of dense Gaussian primitives. Third, we introduce a KL-based Gaussian--image distribution alignment objective to align Gaussian world tokens with foundation image-token distributions. Fourth, we construct a large-scale 3D Gaussian driving scene dataset and a 3DGS-based question-answering dataset to support generalizable 3D Gaussian-based scene understanding and reasoning with LLMs. Fifth, we significantly extend the generation capability from basic spatiotemporal generation to instruction-controllable 4D scene editing, including weather control and dynamic vehicle editing. Finally, we provide more comprehensive experimental evaluations, including additional datasets, stronger baselines, more generation settings, and systematic ablation studies, to further validate the effectiveness and generalization ability of the proposed framework.

\begin{figure*}[!t]
    \centering
    \includegraphics[width=\linewidth]{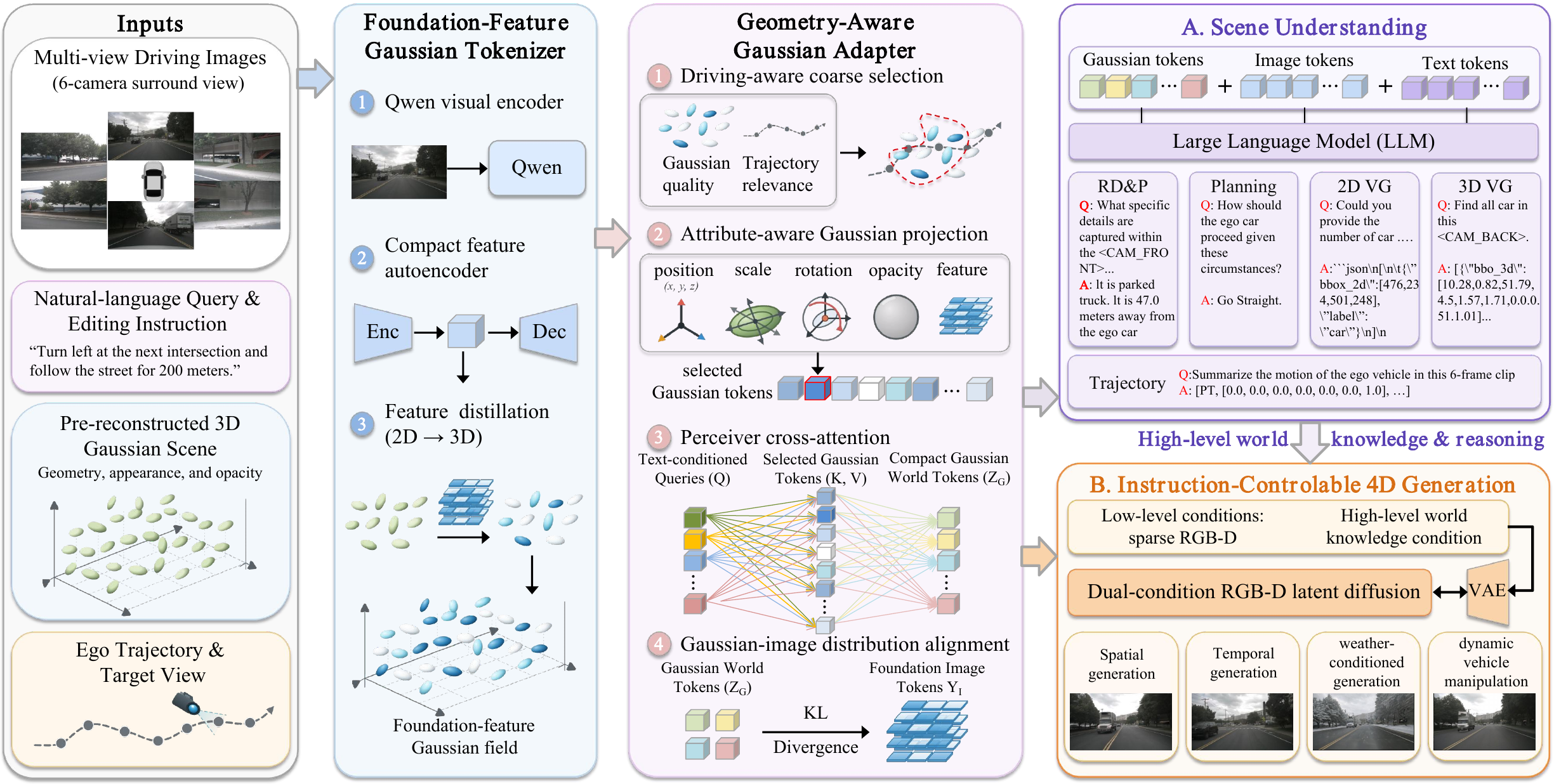}
    \vspace{-0.2cm}
    \caption{\textbf{System Overview.} GaussianDWM++ builds a foundation-feature Gaussian field by distilling Qwen visual features into explicit 3D Gaussian primitives. A geometry-aware adapter first performs driving-aware coarse selection and attribute projection. It then combines a pooled text embedding with learned queries and aggregates the selected Gaussian tokens through Perceiver-style cross-attention. A Gaussian--image distribution alignment objective regularizes the resulting world tokens toward the foundation image-token distribution. The compact Gaussian world tokens are shared by the LLM-based scene understanding branch and the dual-condition scene generation branch for high-level world knowledge.
 }
    \vspace{-0.4cm}
    \label{fig:pipeline}
\end{figure*}

\section{Related Work}

\noindent\textbf{Driving World Models.}
Driving world models~\cite{kong20253d} have attracted increasing attention in autonomous driving due to their ability to model scene evolution, predict future observations, and synthesize controllable simulation data. Existing methods mainly rely on image, video, BEV, occupancy, or point-cloud conditions for scene generation. GAIA-1~\cite{gaia1} introduces an autoregressive framework for driving video generation. DriveDreamer~\cite{drivedreamer} generates realistic driving scenes conditioned on 3D structural information, demonstrating the importance of geometric guidance for autonomous driving generation. MagicDrive~\cite{magicdrive} presents a street-view synthesis framework with precise 3D controls, such as camera poses and road maps, using cross-view attention. DreamDrive~\cite{dreamdrive} further combines video diffusion with hybrid Gaussian scene representations to synthesize 4D driving scenes with 3D-consistent dynamic video rendering. UniScene~\cite{uniscene} proposes an occupancy-centric framework to unify semantic, visual, and LiDAR generation. Recent works further explore controllable generation, future prediction, closed-loop simulation, and 4D world modeling for autonomous driving.

Although these methods have achieved impressive progress in generation quality and controllability, most existing DWMs primarily focus on synthesizing future observations rather than interpreting and reasoning about the driving environment. As a result, they usually lack explicit support for language-grounded querying, 2D/3D visual grounding, planning-oriented reasoning, and instruction-conditioned scene editing. Pioneering efforts such as HERMES~\cite{hermes} and UniFuture~\cite{liang2025seeing} attempt to bridge scene understanding and scene generation by integrating VLMs or language features into driving world models. However, their spatial representations are mainly built upon BEV, depth, or occupancy features, which provide only indirect feature-level alignment between language and 3D space. In contrast, our method constructs a unified DWM upon foundation-feature 3D Gaussians, where visual-language semantics are embedded into explicit 3D primitives and further used for both scene understanding and controllable 4D generation.

\noindent\textbf{Large Language Models for Autonomous Driving.}
Large Language Models (LLMs) and Vision-Language Models (VLMs) have demonstrated strong generalization, semantic understanding, and reasoning capabilities across a wide range of tasks, including scene understanding~\cite{wang2025learning,3dllm,3dvla,zhang2025cross,zhu2025pathology}, visual question answering, visual grounding, and embodied decision-making. In autonomous driving, DriveGPT4~\cite{drivegpt4} processes front-view videos to predict vehicle actions and provide natural language justifications. DriveLM~\cite{drivelm} formulates autonomous driving as graph-based visual question answering and reasoning. NuInteract~\cite{drivemonkey} further integrates large vision-language models with a spatial processor using learnable queries, and provides a large-scale multi-view image-language benchmark containing dense scene captioning, visual grounding, and diverse interactive reasoning tasks. These works show that language supervision and world knowledge can significantly improve the interpretability and reasoning ability of autonomous driving systems.

HERMES~\cite{hermes} is closely related to our work in that it integrates scene representation with a VLM for joint scene understanding and generation, but it still adopts a BEV-based spatial representation. Different from these methods, our approach embeds foundation visual-language features into 3D Gaussian primitives and further aggregates them into compact, language-compatible world tokens, enabling unified 3D-aware reasoning, scene understanding, and multi-modal generation.

\noindent\textbf{Neural Rendering and 4D Urban Scene Reconstruction.}
With the emergence of Neural Radiance Fields (NeRF)~\cite{NeRF} and 3D Gaussian Splatting (3DGS)~\cite{3dgs}, neural scene representations have been widely adopted in novel view synthesis, robotic mapping and localization~\cite{mneslam,mcnslam,plgslam,vpgsslam,deng2025omnimap,deng2024openobj}, virtual and augmented reality~\cite{neslam,li2025human,gu2025mocount,zhu2025sni,gong2025dino,gong2025ov3r}, and autonomous driving~\cite{nsg,snlidar}. For large-scale urban scenes, early NeRF-based methods, such as NSG~\cite{nsg}, SUDS~\cite{suds}, ProSGNeRF~\cite{deng2023prosgnerf}, EmerNeRF~\cite{emernerf}, and FreeDriveRF~\cite{freedriverf}, model dynamic driving environments through scene graphs, dynamic-static decomposition, optical flow, or other motion cues. These methods demonstrate the effectiveness of neural fields for reconstructing complex urban scenes, but their volumetric rendering process often results in high computational costs.

More recently, 3DGS-based methods have been proposed to improve rendering efficiency and scalability. PVG~\cite{pvg} introduces periodic vibration-based temporal dynamics to unify static and dynamic elements without manual annotations. Street Gaussians~\cite{streetgaussian}, Driving Gaussian~\cite{drivinggaussian}, and DeSiRe-GS~\cite{DeSiRe-GS} explicitly separate dynamic foreground objects from static backgrounds for urban scene reconstruction. LESSON~\cite{lesson} explores teacher-guided diffusion for generating 3D Gaussian splats using only 2D supervision. STORM~\cite{storm} proposes a feed-forward Transformer architecture to infer dynamic Gaussians and their velocities, enabling efficient large-scale outdoor scene reconstruction. DrivingForward~\cite{drivingforward} performs feed-forward reconstruction from sparse surround-view inputs using self-supervised pose and depth estimation. MUDG~\cite{mudg} and Dist-4D~\cite{dist4d} further extend urban scene synthesis to multi-modal generation by predicting both RGB and depth observations. Although these methods significantly advance 4D reconstruction and novel view synthesis for urban scenes, they mainly treat NeRF or 3DGS as rendering representations. They generally lack language-grounded scene understanding, explicit reasoning ability, and unified integration with driving world models.

\noindent\textbf{Scene Representations for World Models.}
Scene representation is a fundamental component of autonomous driving systems and driving world models~\cite{deng2025best3dscenerepresentation,tu2025role}. Existing representations include image/video features~\cite{gaia1,vista}, point clouds~\cite{zuo2026dvgt}, BEV maps~\cite{bevformer,hermes}, occupancy fields~\cite{uniscene,zheng2024occworld}, neural radiance fields~\cite{lindstrom2024nerfs}, and 3D Gaussian primitives~\cite{ni2025recondreamer,zhao2025drivedreamer4d,Deng_2026_CVPR}. Image and video features preserve rich appearance information and are naturally compatible with 2D generative models, but they lack explicit 3D structure and metric spatial grounding. Point clouds provide direct geometric measurements, yet they are sparse, irregular, and usually contain limited texture and semantic information. BEV and occupancy representations are highly structured and well suited for detection, prediction, planning, and map-based reasoning, but they discretize the 3D world and may lose fine-grained multi-view appearance and continuous geometry. Neural fields provide continuous scene modeling and differentiable rendering, but they are often computationally expensive for large-scale dynamic driving environments.

Compared with these representations, 3DGS provides an explicit, continuous, and efficient primitive-based scene representation. Each Gaussian primitive carries spatial position, anisotropic geometry, opacity, appearance, and potentially semantic or foundation-model features. This makes 3DGS particularly suitable for driving world modeling, since it can simultaneously support differentiable rendering, multi-view geometric consistency, efficient scene reconstruction, and primitive-level semantic grounding. Recent 3DGS-based urban scene methods~\cite{pvg,streetgaussian,drivinggaussian,DeSiRe-GS,storm,drivingforward} have demonstrated strong reconstruction and rendering efficiency in large-scale outdoor environments. Meanwhile, language- or foundation-feature-based 3D representations, such as LERF~\cite{lerf}, ConceptFusion~\cite{jatavallabhula2023conceptfusion}, OpenScene~\cite{peng2023openscene}, LangSplat~\cite{langsplat}, and GaussianVLM~\cite{gaussianvlm} further show the potential of connecting 3D scene representations with visual-language foundation models for understanding and language-grounded interaction.

\section{Method}
\label{sec:method}

\subsection{Overview}
\label{sec:method_overview}

We propose \textbf{GaussianDWM++}, a foundation-feature 4D Gaussian driving world model for unified scene understanding, language-grounded reasoning, controllable 4D editing, and multi-modal scene generation. The input to our method consists of images $\mathcal{I}=\{I_v\}_{v=1}^V$, Gaussian ellipsoids $\mathcal{G}=\{G_i\}_{i=1}^N$, and query text $q$. Our goal is to construct a compact and semantically aligned 3D world representation that can be consumed by both large language models and generative models. The overall pipeline is illustrated in Fig.~\ref{fig:pipeline}.

Our framework consists of four main components. First, a foundation-feature Gaussian tokenizer distills Qwen visual-language features into 3D Gaussian primitives, producing an open-vocabulary semantic field with explicit 3D alignment (Sec.~\ref{sec:foundation_tokenizer}). Second, a geometry-aware adapter combines hierarchical selection, text-conditioned Perceiver-style cross-attention, and Gaussian--image distribution alignment to produce compact world tokens $\mathcal{Z}_{G}$ (Sec.~\ref{sec:gaussian_adapter}). These tokens are injected into the LLM with image and text tokens for scene understanding (Sec.~\ref{sec:understanding}) and are shared with the generation branch for spatiotemporal generation and instruction-controllable 4D editing (Sec.~\ref{sec:generation}).

\subsection{Foundation-feature Gaussian Tokenizer}
\label{sec:foundation_tokenizer}

The foundation-feature Gaussian tokenizer converts multi-view driving observations into an explicit 3D Gaussian semantic field. Different from the conference version that relies on pre-computed LangSplat features, we directly distill Qwen visual-language features into 3D Gaussian primitives. In this way, each Gaussian primitive is not only used for differentiable rendering, but also serves as an explicit 3D carrier of open-vocabulary foundation features.

\noindent\textbf{3D Gaussian Scene Representation.}
Given a reconstructed driving scene, we represent the Gaussian field as
\begin{equation}
\mathcal{G}=\left\{G_i=\left(x_i,s_i,r_i,o_i,c_i,z_i\right)\right\}_{i=1}^{N}.
\end{equation}
Here, $x_i\in\mathbb{R}^{3}$ is the Gaussian center, $s_i\in\mathbb{R}^{3}$ is the anisotropic scale, $r_i\in\mathbb{R}^{4}$ is the rotation quaternion, $o_i\in\mathbb{R}$ is the opacity, $c_i$ denotes the appearance attribute, and $z_i\in\mathbb{R}^{d_z}$ denotes the compact foundation-feature code. For each timestamp, we transform Gaussian centers into the current ego coordinate system, so that the Gaussian field is represented under a driving-centric coordinate frame. The opacity and scale are normalized to avoid unstable primitives during feature rendering and token extraction.

\noindent\textbf{Foundation Feature Extraction.}
For each multi-view image $I_v$, we use a frozen foundation visual encoder to extract dense visual-language features:
\begin{equation}
F_v=\mathcal{E}_F(I_v).
\end{equation}
Here, $F_v\in\mathbb{R}^{H'\times W'\times d_f}$ denotes the dense foundation feature map, and $d_f$ is the feature dimension. In practice, $\mathcal{E}_F$ is instantiated with Qwen visual encoders. These frozen features provide semantically rich supervision for constructing the 3D Gaussian semantic field.

\noindent\textbf{Foundation-Feature Distillation.}
To lift 2D foundation features into the 3D Gaussian field, we associate a compact feature code $z_i$ with each Gaussian primitive. Since the original foundation feature dimension is high, we decode the compact code into a full-dimensional feature before rendering:
\begin{equation}
\hat{f}_i=\mathcal{D}_F(z_i).
\end{equation}
The decoded feature $\hat{f}_i\in\mathbb{R}^{d_f}$ is then rendered to each camera view using the same Gaussian splatting process as RGB rendering. For a pixel $u$ in camera view $v$, the rendered foundation feature is:
\begin{equation}
\hat{F}_v(u)=\sum_{i\in\mathcal{N}_v(u)}\omega_{i,u}^{v}\hat{f}_i.
\end{equation}
Here, $\mathcal{N}_v(u)$ denotes the Gaussians contributing to pixel $u$, $\omega_{i,u}^{v}$ is the corresponding alpha-compositing weight, and $\Omega_v$ is the set of valid pixels in view $v$. We supervise the rendered features with a pixel-level L1 distillation loss:
\begin{equation}
\mathcal{L}_{\mathrm{distill}}=
\frac{1}{\sum_{v=1}^{V}|\Omega_v|}
\sum_{v=1}^{V}\sum_{u\in\Omega_v}
\left\lVert\hat{F}_v(u)-F_v(u)\right\rVert_1.
\end{equation}
This objective lifts foundation visual-language features from 2D observations into the explicit 3D Gaussian field, enabling primitive-level semantic grounding.

\noindent\textbf{Compact Feature Autoencoding.}
To reduce storage cost, we further adopt a compact feature autoencoding strategy. Given the high-dimensional foundation feature $f_i$, we encode it into a low-dimensional latent code and reconstruct it through a lightweight decoder:
\begin{equation}
z_i=\mathcal{E}_z(f_i),\quad \tilde{f}_i=\mathcal{D}_F(z_i).
\end{equation}
The reconstruction objective is a squared L2 loss:
\begin{equation}
\mathcal{L}_{\mathrm{ae}}=\frac{1}{N}\sum_{i=1}^{N}
\left\lVert\tilde{f}_i-f_i\right\rVert_2^2.
\end{equation}
Only the compact code $z_i$ is stored in each Gaussian primitive. This design keeps the Gaussian semantic field compact while preserving foundation-level visual-language semantics.

The resulting foundation-feature Gaussian field explicitly couples 3D geometry, appearance, opacity, and visual-language features. The adapter in Sec.~\ref{sec:gaussian_adapter} then selects and aggregates these primitives into compact world tokens.

\subsection{Geometry-aware Gaussian Adapter}
\label{sec:gaussian_adapter}

The foundation-feature Gaussian field in Sec.~\ref{sec:foundation_tokenizer} contains explicit geometry, appearance, opacity, and foundation visual-language features. However, a driving scene usually contains tens or hundreds of thousands of Gaussian primitives, which cannot be directly consumed by an LLM. We therefore introduce a geometry-aware Gaussian adapter to convert the dense Gaussian field into a compact set of LLM-compatible world tokens. The adapter consists of three components: driving-aware coarse selection, attribute-aware Gaussian projection, and text-conditioned Perceiver-style cross-attention with Gaussian--image distribution alignment.

\noindent\textbf{Driving-aware Coarse Selection.}
Given the foundation-feature Gaussian field, the coarse selector aims to preserve a compact yet spatially representative candidate set. Different from the previous distance-biased quality score, we introduce a driving-aware importance score that jointly considers intrinsic Gaussian quality and trajectory relevance. We first define the intrinsic quality of each Gaussian as
\begin{equation}
Q_i=\operatorname{Norm}_{[0,1]}\left(\log\left(1+\sigma\left(o_i\right)\left(\left\lVert s_i\right\rVert_1+1\right)\rho_i\right)\right).
\end{equation}
The rotation validity term $\rho_i$ is defined as
\begin{equation}
\rho_i=\frac{1}{1+\left|\left\lVert r_i\right\rVert_2-1\right|}.
\end{equation}
Here, $\operatorname{Norm}_{[0,1]}$ denotes min-max normalization within the current scene. The quality term favors Gaussians with high opacity, reasonable scale, and valid rotation, while avoiding an explicit bias toward the ego origin.

To encourage the selected Gaussians to cover regions that are important for future driving, we further define a trajectory relevance term. Let the ego trajectory be denoted as $\mathcal{P}_{\mathrm{traj}}=\{p_j\}_{j=0}^{T}$, where $p_j\in\mathbb{R}^{2}$ is a waypoint on the ground plane. The distance from a Gaussian to the future trajectory is
\begin{equation}
d_i^{\mathrm{traj}}=\min_{p_j\in\mathcal{P}_{\mathrm{traj}}}
\left\lVert\pi_{xy}\left(x_i\right)-p_j\right\rVert_2.
\end{equation}
The trajectory relevance score is then computed as
\begin{equation}
T_i=\exp\left(-\frac{\left(d_i^{\mathrm{traj}}\right)^2}{2\sigma_{\mathrm{traj}}^2}\right).
\end{equation}
Here, $\pi_{xy}\left(\cdot\right)$ projects a 3D point onto the ground plane, and $\sigma_{\mathrm{traj}}$ controls the spatial influence range of the trajectory prior.

The final driving-aware score is
\begin{equation}
S_i=\frac{w_qQ_i+w_tT_i}{w_q+w_t}.
\end{equation}
In our implementation, $w_q$ and $w_t$ control the contributions of intrinsic quality and trajectory relevance, respectively. We select $K_c$ candidates by combining a global top-$K$ branch and a voxel-coverage branch:
\begin{equation}
\mathcal{C}=\operatorname{TopK}\left(\{S_i\}_{i=1}^{N},K_c\right)\cup\operatorname{VoxelTopK}\left(\{S_i\}_{i=1}^{N},K_c\right).
\end{equation}
The global branch preserves the most informative Gaussians, while the voxel branch enforces spatial coverage over the whole driving scene. If the union contains more than $K_c$ candidates, we keep the top-ranked candidates according to $S_i$; otherwise, we pad the candidate set with invalid tokens.

\noindent\textbf{Attribute-aware Gaussian Projection.}
After coarse selection, each candidate Gaussian is projected into the LLM hidden space by an attribute-aware aligner. For the 3D position, we adopt a learnable Fourier embedding:
\begin{equation}
\phi_{\mathrm{pos}}\left(x_i\right)=\left[\sin\left(2\pi Bx_i\right),\cos\left(2\pi Bx_i\right)\right].
\end{equation}
Here, $B$ is a learnable linear projection. We decode $z_i$ as $\hat{f}_i=\mathcal{D}_{F}(z_i)$ and project position, scale, rotation, opacity, and foundation features into a shared hidden space. Denoting this attribute set as $\mathcal{P}_{\mathrm{attr}}=\{x,s,r,o,\hat{f}\}$, we fuse the embeddings using learnable scalar gates:
\begin{equation}
\tilde{h}_i=\sum_{p\in\mathcal{P}_{\mathrm{attr}}}\lambda_ph_i^{p}.
\end{equation}
The fused embedding is further refined by a residual MLP aligner:
\begin{equation}
g_i=\tilde{h}_i+\operatorname{MLP}\left(\operatorname{LN}\left(\tilde{h}_i\right)\right).
\end{equation}
The resulting embedding $g_i$ is the aligned Gaussian token representation for the $i$-th candidate.

\noindent\textbf{Text-Conditioned Perceiver-Style Cross-Attention.}
Although the coarse selector significantly reduces the number of Gaussians, the candidate set $\mathcal{C}$ is still too large for direct LLM injection. We therefore use a small set of learned queries to aggregate the candidate Gaussian tokens through cross-attention. Given the input text embeddings, we first compute a pooled instruction embedding:
\begin{equation}
\bar{e}_q=\frac{1}{|\mathcal{T}|}\sum_{\ell\in\mathcal{T}}e_{\ell},
\end{equation}
where $\mathcal{T}$ denotes the valid text-token positions and $e_{\ell}$ is the embedding of the $\ell$-th token. Let $\{q_j^{\mathrm{L}}\}_{j=1}^{K_f}$ denote $K_f$ learned queries. Each query is modulated by the pooled text embedding:
\begin{equation}
\tilde{q}_j=q_j^{\mathrm{L}}+W_t\bar{e}_q,
\qquad j=1,\ldots,K_f.
\end{equation}
The aligned Gaussian candidates provide the keys and values:
\begin{equation}
\mathcal{K}_G=\left\{W_k g_i\mid i\in\mathcal{C}\right\},
\qquad
\mathcal{V}_G=\left\{W_v g_i\mid i\in\mathcal{C}\right\}.
\end{equation}
We then aggregate the Gaussian candidates with multi-head cross-attention:
\begin{equation}
z_j=\operatorname{MHA}\left(W_q\tilde{q}_j,\mathcal{K}_G,\mathcal{V}_G\right),
\qquad j=1,\ldots,K_f.
\end{equation}
The resulting query responses form the compact Gaussian world-token set:
\begin{equation}
\mathcal{Z}_{G}=\left\{z_j\right\}_{j=1}^{K_f}.
\end{equation}
Because the learned queries are modulated by the pooled instruction embedding, the adapter preserves a fixed token budget while aggregating Gaussian evidence that is relevant to the current query.

\noindent\textbf{Gaussian--Image Distribution Alignment.}
To make the aggregated Gaussian world tokens more compatible with the foundation visual space used by the LLM, we introduce a KL-based distribution alignment objective. Let $\mathcal{Y}_I=\{y_m^I\}_{m=1}^{M_I}$ denote the foundation image tokens from the same sample. We estimate the diagonal Gaussian statistics of the Gaussian world tokens as
\begin{equation}
\mu_G=\frac{1}{K_f}\sum_{z\in\mathcal{Z}_{G}}z,
\qquad
\sigma_G^2=\frac{1}{K_f}\sum_{z\in\mathcal{Z}_{G}}\left(z-\mu_G\right)^2.
\end{equation}
The target image-token statistics are computed with stop-gradient:
\begin{equation}
\mu_I=\operatorname{sg}\left(\frac{1}{M_I}\sum_{m=1}^{M_I}y_m^I\right).
\end{equation}
\begin{equation}
\sigma_I^2=\operatorname{sg}\left(\frac{1}{M_I}\sum_{m=1}^{M_I}\left(y_m^I-\mu_I\right)^2\right).
\end{equation}
Let $\mathcal{N}_G=\mathcal{N}(\mu_G,\operatorname{diag}(\sigma_G^2+\epsilon))$ and $\mathcal{N}_I=\mathcal{N}(\mu_I,\operatorname{diag}(\sigma_I^2+\epsilon))$. The dimension-normalized alignment loss is
\begin{equation}
\mathcal{L}_{\mathrm{KL}}=\frac{1}{d}D_{\mathrm{KL}}\left(\mathcal{N}_G\,\|\,\mathcal{N}_I\right).
\end{equation}
During task training, $\mathcal{L}_{\mathrm{KL}}$ is added to the language modeling objective with weight $\lambda_{\mathrm{KL}}$ and warm-up coefficient $\alpha(t)$. It regularizes Gaussian world tokens toward the foundation image-token distribution without imposing a separate KL constraint on text tokens.

\subsection{Scene Understanding with Gaussian World Tokens}
\label{sec:understanding}

After obtaining the compact Gaussian world tokens from the geometry-aware Gaussian adapter in Sec.~\ref{sec:gaussian_adapter}, we inject them into the LLM together with multi-view image tokens and text tokens for 3D-aware scene understanding. 

\noindent\textbf{Input Token Construction.}
Given multi-view images, we first encode them into image tokens by the visual encoder of the VLM. The language instruction $q$ is tokenized into text tokens. The compact Gaussian world tokens are produced by the geometry-aware Gaussian adapter.
Then, we insert $\mathcal{Z}_{G}$ into dedicated Gaussian token slots between the image tokens and the text instruction. The final input sequence to the LLM is defined as
\begin{equation}
\mathcal{X}_{\mathrm{in}}=\operatorname{Concat}\left(\mathcal{V},\mathcal{Z}_{G},\mathcal{T}\right).
\end{equation}
This construction allows the LLM to jointly attend to appearance information from image tokens, explicit 3D spatial information from Gaussian tokens, and task semantics from text tokens.

\noindent\textbf{Unified Scene Understanding Formulation.}
We formulate all scene understanding tasks as conditional language generation, with the LLM autoregressively predicting response $a$ from $\mathcal{X}_{\mathrm{in}}$. Region description uses natural language, 2D grounding uses image-plane boxes, 3D grounding uses object location and size in the ego frame, and planning outputs a decision with its justification. Thus, heterogeneous tasks share the same token format and language modeling objective.

\noindent\textbf{Language Modeling Objective.}
During training, each sample is converted into a prefix-response pair. The prefix contains the image tokens, Gaussian world tokens, and instruction tokens, while the response contains the ground-truth answer. We optimize the model with a prefix language modeling loss:
\begin{equation}
\mathcal{L}_{\mathrm{LM}}=-\frac{1}{|\mathcal{B}|}\sum_{\left(q,a\right)\in\mathcal{B}}\sum_{j=1}^{|a|}\log p_{\theta}\left(a_j\mid a_{<j},\mathcal{X}_{\mathrm{in}}\right).
\end{equation}
Here, $\mathcal{B}$ denotes a training batch, $a_j$ denotes the $j$-th response token, and $a_{<j}$ denotes all previous response tokens. This objective trains the LLM to generate task-specific answers conditioned on multi-view image observations, compact Gaussian world tokens, and language instructions.

\noindent\textbf{Training Strategy.}
We train the scene understanding module in two stages. In the first stage, we freeze the LLM and optimize the foundation-feature Gaussian tokenizer and geometry-aware adapter, so that the compact Gaussian world tokens become compatible with the LLM hidden space. In the second stage, we keep the Gaussian tokenizer and adapter trainable and enable LoRA adaptation on the LLM, allowing the model to better integrate image, language, and Gaussian tokens for downstream reasoning. The total loss for scene understanding is defined as
\begin{equation}
\mathcal{L}_{\mathrm{understanding}}=\mathcal{L}_{\mathrm{LM}}+\lambda_{\mathrm{KL}}\alpha\left(t\right)\mathcal{L}_{\mathrm{KL}}.
\end{equation}
Here, $\mathcal{L}_{\mathrm{KL}}$ is the distribution alignment loss introduced in Sec.~\ref{sec:gaussian_adapter}, $\lambda_{\mathrm{KL}}$ controls its weight, and $\alpha\left(t\right)$ is a warm-up coefficient. To reduce the influence of task imbalance, we further adopt task-aware batch sampling. Let $\mathcal{Q}$ denote the set of training tasks and $n_m$ denote the number of samples in task $m$. The sampling probability of task $m$ is defined as
\begin{equation}
P_m=\frac{n_m^{\alpha_s}}{\sum_{k\in\mathcal{Q}}n_k^{\alpha_s}}.
\end{equation}
Here, $\alpha_s$ is a smoothing factor. This strategy prevents large tasks from dominating the training process and improves the stability of multi-task scene understanding.

\subsection{Instruction-controllable 4D Scene Generation}
\label{sec:generation}

In this section, we introduce the generation branch of GaussianDWM++, which performs multi-modal spatiotemporal scene generation and instruction-controllable 4D editing. Different from conventional driving scene generators that mainly rely on image or BEV conditions, our generation module is conditioned on both low-level geometric observations and high-level world knowledge extracted from the Gaussian-aware LLM. The low-level conditions preserve scene texture and geometry, while the high-level language condition provides semantic guidance, driving context, and editing intent.

\noindent\textbf{Dual-condition RGB-D Latent Diffusion.}
Our generation module is built upon a denoising UNet~\cite{stablediffusion} and a frozen pre-trained VAE~\cite{vae}. Given an RGB image $I_i$ and its corresponding depth map $D_i$, we first convert the depth map into a pseudo-RGB image by channel replication.
The RGB image and the pseudo-RGB depth image are then encoded into a unified latent space:
\begin{equation}
z_I=\mathcal{E}_{\mathrm{vae}}\left(I_i\right), \quad
z_D=\mathcal{E}_{\mathrm{vae}}\left(D_i^{\mathrm{rgb}}\right).
\end{equation}
During decoding, the VAE decoder reconstructs the RGB image and the depth map from the generated latent variables:
\begin{equation}
\hat{I}_i=\mathcal{D}_{\mathrm{vae}}\left(z_I\right), \quad
\hat{D}_i=\frac{1}{3}\sum_{c=1}^{3}\mathcal{D}_{\mathrm{vae}}\left(z_D\right)_c.
\end{equation}
Here, the three decoded channels of the depth branch are averaged to obtain the final single-channel depth prediction.

For each training sample, we randomly sample a target camera state from either the spatially shifted trajectory or the future ego trajectory. The latent variable of the target modality is denoted as $d$. At diffusion timestep $t$, we add Gaussian noise to obtain the noisy latent:
\begin{equation}
d_t=\alpha_t d+\sigma_t\epsilon.
\end{equation}
Here, $\epsilon\sim\mathcal{N}\left(0,I\right)$ denotes Gaussian noise, and $\alpha_t$ and $\sigma_t$ are the noise scheduling coefficients. Following the v-prediction objective, the denoising target is defined as:
\begin{equation}
v_t=\alpha_t\epsilon-\sigma_t d.
\end{equation}

\noindent\textbf{Low-level Conditions.}
To provide explicit geometric guidance for generation, we construct low-level RGB and depth condition maps from the observed Gaussian scene or projected point cloud. Given the target camera transformation, we project the surrounding 3D scene evidence into the target view and obtain sparse RGB and depth conditions. The condition $C_I$ provides sparse appearance cues, while $C_D$ provides geometric structure. These low-level conditions help the diffusion model preserve spatial layout and geometric consistency during novel view synthesis and future scene prediction.

\noindent\textbf{High-level Language Conditions.}
In addition to low-level conditions, the generation branch receives high-level semantic guidance from the scene understanding module. Given $\mathcal{Z}_{G}$ and instruction $q$, the LLM produces the language condition $C_L=\Phi_{\mathrm{LLM}}(\mathcal{Z}_{G},q)$.

\noindent\textbf{Diffusion Training Objective.}
The denoising network takes the noisy latent, reference latent, low-level conditions, high-level language condition, and diffusion timestep embedding as input. Editing intent, when present, is encoded as part of the instruction-dependent condition $C_L$:
\begin{equation}
\hat{v}_t=\mathcal{F}_{\theta}\left(d_t,d_{\mathrm{ref}},C_I,C_D,C_L,s_t\right).
\end{equation}
Here, $d_{\mathrm{ref}}$ denotes the reference latent, and $s_t$ denotes the timestep embedding. The generation loss is defined as:
\begin{equation}
\mathcal{L}_{\mathrm{gen}}=\mathbb{E}_{d,\epsilon,t,s_t}\left\lVert\mathcal{F}_{\theta}\left(d_t,d_{\mathrm{ref}},C_I,C_D,C_L,s_t\right)-v_t\right\rVert_2^2.
\end{equation}
By jointly conditioning on $C_I$, $C_D$, and $C_L$, the diffusion model learns to generate RGB-D observations that are geometrically consistent, semantically coherent, and controllable by language instructions.

\noindent\textbf{Spatiotemporal Generation and 4D Editing.}
For spatial generation, we obtain the target transformation by applying a lateral or longitudinal shift to the current camera pose, and construct sparse condition maps by projecting the Gaussian scene into the target view. This enables novel view synthesis under spatial shifts such as $1\mathrm{m}$ or $2\mathrm{m}$. For temporal generation, we use the future ego trajectory predicted or inferred by the scene understanding module to define the target camera state, and project the Gaussian scene evidence into future timestamps. This enables future RGB-D prediction at $1\mathrm{s}$ or $2\mathrm{s}$. Beyond spatial and temporal generation, GaussianDWM++ supports instruction-controllable 4D scene editing. For weather-conditioned generation, the editing condition modifies the global appearance and visibility while preserving the underlying scene geometry. For dynamic vehicle editing, it controls object-level operations such as inserting, removing, stopping, or moving vehicles along specified trajectories. Since both low-level geometric conditions and high-level language conditions are derived from the same foundation-feature Gaussian world representation, the generated results maintain better temporal coherence, geometric consistency, and semantic faithfulness across 4D driving scenes.

\section{Experiments}

\subsection{Datasets and Evaluation Metrics}
\label{sec:datasets_metrics}

We evaluate GaussianDWM++ on both scene understanding and multi-modal scene generation tasks. For public driving data, we use nuScenes~\cite{nuscenes}, a large-scale autonomous driving dataset with synchronized multi-view cameras, LiDAR, ego poses, object annotations, and temporal sequences. Following common autonomous driving settings, we use six surrounding camera images as visual inputs and construct 3D Gaussian scene representations from the corresponding driving sequences. For language-grounded scene understanding, we conduct experiments on three complementary benchmarks: NuInteract, NuInstruct~\cite{ding2024holistic}, and OmniDrive~\cite{omnidrive}. NuInteract~\cite{drivemonkey} provides approximately 1.5M multi-view image--language annotations and covers region description and perception, 2D visual grounding, 3D visual grounding, and planning-oriented reasoning. NuInstruct focuses on cross-view geometric understanding through tasks such as risk-object perception, target-distance estimation, agent and ego-state prediction, target-motion prediction, and reasoning. OmniDrive evaluates holistic driving-scene understanding with scene descriptions, attention-based question answering, counterfactual reasoning, decision and planning questions, and general dialogue. Following the VGGDrive evaluation protocol, we use its scene-description and general-dialogue tasks for quantitative evaluation. In addition to these public datasets, we further construct a foundation-feature 3D Gaussian driving dataset by distilling Qwen features into reconstructed Gaussian primitives. The construction details are introduced in Sec.~\ref{sec:dataset_construction}.

For scene understanding, we follow the task-specific evaluation protocols of the three benchmarks. On NuInteract, region description and perception are evaluated using BLEU~\cite{bleu}, ROUGE-L~\cite{rouge}, and CIDEr~\cite{cider}; 2D visual grounding uses mAP, F1 score, and mIoU; 3D visual grounding uses precision, mAP, and F1 score; and planning is evaluated by decision accuracy. We report the arithmetic mean over these ten metrics as the overall NuInteract score. On NuInstruct, we use MAE for regression tasks, accuracy for classification tasks, mAP for risk-object perception and detection tasks, and BLEU for captioning tasks, together with the protocol's aggregate Average* score. On OmniDrive, we evaluate scene-description and general-dialogue outputs using BLEU, CIDEr, and ROUGE-L, and report their arithmetic mean as the average score. The corresponding results are reported in Tabs.~\ref{tab:vqa-main-nuinteract}--\ref{tab:vqa-main-omnidrive}.

For scene generation, we evaluate both spatial and temporal generation. Spatial generation corresponds to novel view synthesis under shifted camera poses, while temporal generation corresponds to future scene prediction along the driving trajectory. We report FID for image generation quality and FVD for temporal video generation quality. For RGB-D generation, we additionally evaluate the visual quality of RGB outputs and the geometric consistency of depth predictions when ground-truth depth is available. For instruction-controllable 4D editing, including weather-conditioned generation and dynamic vehicle manipulation, we provide qualitative visualizations to assess semantic consistency, temporal coherence, and controllability.

\subsection{Implementation Details}
\label{sec:implementation}

Our training pipeline consists of three stages. In the first stage, we train the foundation-feature Gaussian tokenizer and the geometry-aware Gaussian adapter. The Gaussian tokenizer distills Qwen visual-language features into 3D Gaussian primitives using the L1 feature-distillation loss, while the compact feature autoencoder is optimized with the squared L2 reconstruction loss. In the second stage, we integrate the Gaussian world tokens with the LLM for scene understanding. We freeze the main LLM parameters at the beginning and optimize the Gaussian tokenizer, attribute projection, and cross-attention adapter to warm up Gaussian--image representation alignment. Then, we enable LoRA-based fine-tuning of the LLM and jointly optimize the Gaussian adapter and the language reasoning branch. In the third stage, we train the multi-modal scene generation branch. The generation model is built on a denoising UNet~\cite{stablediffusion} and a frozen pre-trained VAE~\cite{vae}. It is optimized with a v-prediction objective~\cite{vprediction}. The low-level conditions provide texture and geometric guidance, while the high-level language condition provides scene-level world knowledge and editing instructions. All models are trained on 16 NVIDIA A100 GPUs. During inference, the same compact Gaussian world representation is shared by the scene understanding branch and the generation branch, enabling consistent 3D-aware reasoning and controllable 4D scene generation.

\subsection{NuScenes-GSQA Dataset Construction}
\label{sec:dataset_construction}

To support 3D Gaussian-based scene understanding and controllable driving scene generation, we construct a large-scale foundation-feature 3D Gaussian dataset from nuScenes~\cite{nuscenes}. The dataset contains 1000 outdoor urban driving scenes reconstructed as 3D Gaussians. Different from conventional image- or BEV-based driving datasets, our dataset provides explicit 3D Gaussian scene representations with geometry, appearance, opacity, and foundation visual-language features, enabling LLMs to reason over structured 3D scene primitives. The constructed dataset serves as the basis for Gaussian-aware scene understanding.

\noindent\textbf{Large-scale 3D Gaussian Reconstruction.}
We reconstruct each nuScenes scene as a 3D Gaussian field using synchronized multi-view images, LiDAR point clouds, ego poses, and camera calibration parameters. Since outdoor driving scenes are large-scale and often contain repetitive structures, dynamic objects, motion blur, and limited view overlap, directly applying conventional COLMAP-based 3DGS initialization can lead to unstable or inaccurate reconstructions. To improve geometric robustness, we initialize each scene with aggregated LiDAR points provided by nuScenes. The aggregated point cloud is transformed into a unified scene coordinate frame and used to initialize Gaussian centers, while multi-view images are used to optimize appearance, opacity, scale, and rotation. During reconstruction, we remove frames with severe motion blur, poor exposure, or extremely low cross-view overlap. For scenes with reliable depth supervision, we additionally introduce a depth constraint to improve geometric fidelity. After optimization, each reconstructed Gaussian scene is evaluated using rendering metrics such as PSNR together with visual inspection. We exclude unreliable frames and primitives identified during this quality-control process while retaining the complete set of 1000 reconstructed scenes for downstream pre-training, scene understanding, and generation.

\noindent\textbf{Foundation-feature Gaussian Annotation.}
After obtaining high-quality 3D Gaussian reconstructions, we further inject foundation visual-language features into Gaussian primitives. Specifically, we extract dense Qwen features from multi-view images and distill them into the reconstructed Gaussian field following Sec.~\ref{sec:foundation_tokenizer}. Each Gaussian primitive stores a compact foundation-feature code rather than the full-dimensional feature, reducing storage cost while preserving open-vocabulary semantic information. The resulting representation associates visual-language semantics with explicit 3D scene primitives, providing a unified carrier for geometry, appearance, and foundation features.

\noindent\textbf{Gaussian-aware QA Dataset.}
To train the LLM to consume 3D Gaussian tokens, we construct a Gaussian-aware QA dataset based on NuInteract~\cite{drivemonkey}. We convert the original image-language annotations into a format that explicitly includes Gaussian references. Following the training paradigm of Qwen-VL, the Gaussian token placeholder is inserted at the beginning of the user instruction. In this way, the model learns to treat Gaussian tokens as structured 3D context before answering scene understanding questions.

A simplified annotation example is shown below:
\begin{lstlisting}[language=json]
{
"scene_token": "e7ef871f77f44331aefdebc24ec034b7",
"frame_idx": 0,
"task": "rdp",
"category": "2D_perception_infos",
"conversations": [
{
"from": "human",
"value": "Based on <gauss>, can you estimate the distance between the <CAM_FRONT> <box>(14,219),(37,248)</box> and the ego vehicle?"
},
{
"from": "gpt",
"value": "The object is approximately 38.7 meters away from the ego vehicle."
}
],
"image": [
"nuscenes/samples/CAM_FRONT/xxx.jpg"
],
"views": [
"CAM_FRONT"
],
"gauss": [
"gauss/scene_0002/frame_00000.pth"
]
}
\end{lstlisting}

The QA annotations cover multiple scene understanding tasks, including region description and perception, 2D visual grounding, 3D visual grounding, and planning-oriented reasoning. Compared with image-only QA data, the Gaussian-aware annotations explicitly bind each instruction to the corresponding 3D Gaussian scene representation, allowing the model to learn 3D-aware language grounding.

\noindent\textbf{Trajectory-oriented QA Dataset.}
For spatial and temporal scene generation, the model needs to infer target camera poses from language instructions. We therefore construct a trajectory-oriented QA dataset to enhance trajectory understanding and future pose prediction. For spatial generation, instructions typically describe a target viewpoint, such as shifting the camera by a certain distance to the left, right, forward, or backward. The model is trained to infer the corresponding target pose, which is then used to project the Gaussian scene or point cloud into the target view to obtain sparse generation conditions.

For temporal generation, instructions describe future scene prediction, such as predicting the scene one or two seconds later. We extract 10-frame video clips from nuScenes sequences. The first 4 frames are used as historical trajectory inputs, and the remaining 6 frames are used as prediction targets. All poses are transformed into the ego coordinate system of the 5th frame, so that the model learns future motion in a normalized local frame. A simplified trajectory QA example is shown below:
\begin{lstlisting}[language=json]
{
"conversations": [
{
"from": "human",
"value": "There are last 4 frames of ego trajectory: [PT, [-7.45, 3.05, 0.16, 0.00, 0.00, -0.36, 0.93], [-5.75, 1.97, 0.11, 0.00, 0.00, -0.28, 0.96], [-4.04, 0.92, 0.09, 0.00, 0.00, -0.19, 0.98], [-2.15, 0.26, 0.04, 0.00, 0.00, -0.10, 1.00]]. Predict the future 6-frame ego trajectory."
},
{
"from": "gpt",
"value": "[PT, [0.00, 0.00, 0.00, 0.00, 0.00, 0.00, 1.00], [2.47, 0.65, -0.06, 0.00, 0.00, 0.09, 1.00], [4.24, 2.32, -0.11, 0.00, 0.00, 0.12, 0.99], [5.50, 4.08, -0.15, 0.00, 0.00, 0.13, 0.99], [7.03, 6.45, -0.15, 0.00, 0.00, 0.13, 0.99], [8.66, 8.91, -0.17, -0.01, 0.00, 0.13, 0.99]]"
}
]
}
\end{lstlisting}

This trajectory-oriented supervision enables the LLM to predict target camera states for spatial and temporal generation. The predicted pose is then used to construct sparse RGB and depth conditions for the diffusion generator. As a result, the same dataset supports both Gaussian-aware scene understanding and trajectory-conditioned 4D scene generation.

\begin{figure*}[!t]
    \centering
    \includegraphics[width=\linewidth]{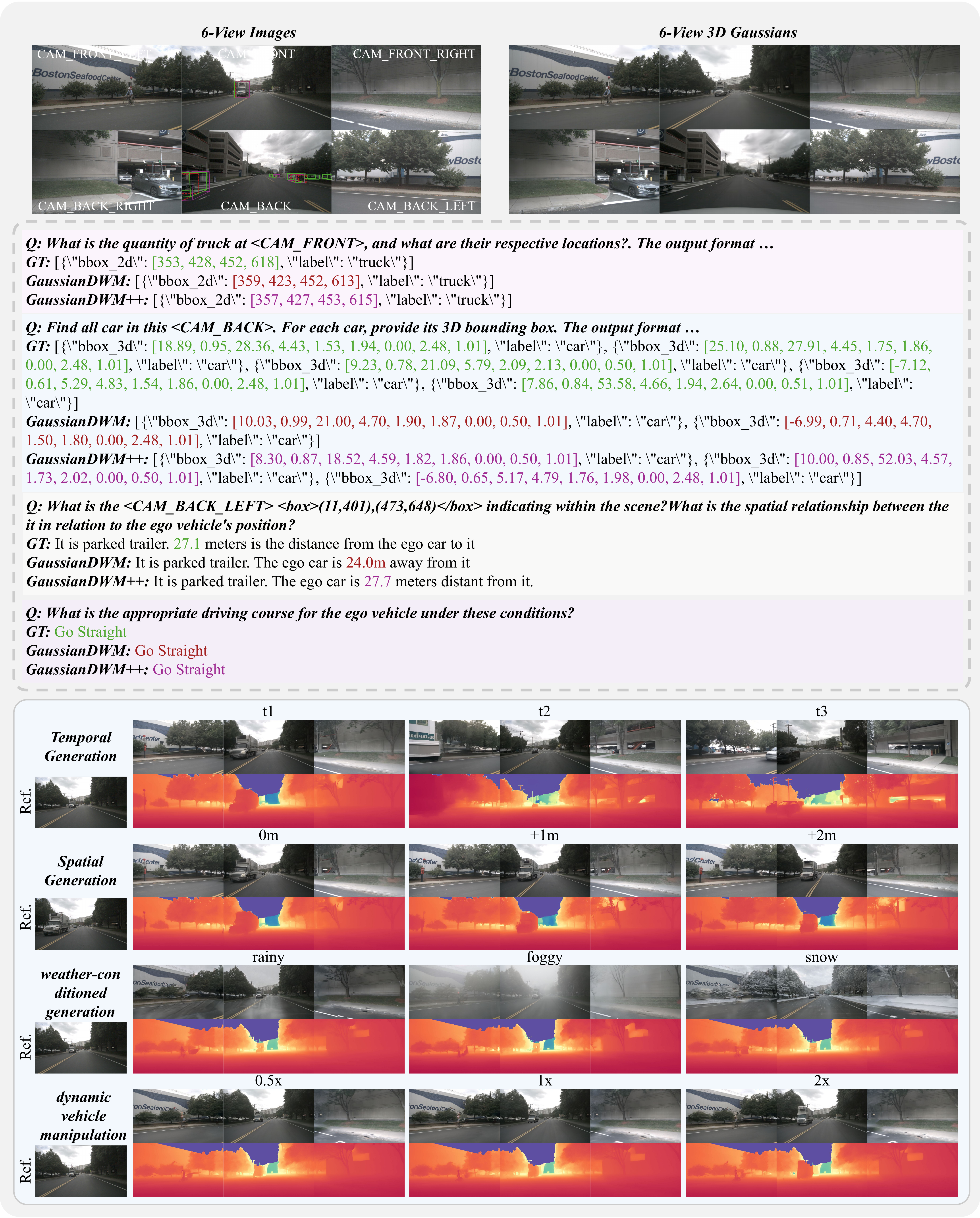}
    \vspace{-0.2cm}
    \caption{\textbf{Qualitative results for scene understanding and scene generation.} From top to bottom, we display the multi-view input of the current scene and the 3D Gaussian ellipsoids, the scene understanding results, and the results of temporal generation, spatial generation, weather-conditioned generation, and dynamic vehicle manipulation with varying motion speeds.
 }
    \vspace{-0.4cm}
    \label{fig:experiments}
\end{figure*}

\subsection{Scene Understanding}
We evaluate scene understanding on NuInteract, NuInstruct, and OmniDrive, which together cover language description, cross-view perception, 2D/3D visual grounding, state prediction, and planning-oriented reasoning. The quantitative comparisons are reported in Tabs.~\ref{tab:vqa-main-nuinteract}--\ref{tab:vqa-main-omnidrive}, and representative qualitative results are shown in Fig.~\ref{fig:experiments}.

\noindent\textbf{NuInteract.}
Following the DriveMonkey protocol~\cite{drivemonkey}, we compare GaussianDWM++ with general-purpose LVLMs, driving-specific VLMs, and specialized 3D detectors. As shown in Tab.~\ref{tab:vqa-main-nuinteract}, GaussianDWM++ achieves the best overall result, ranks first on all region-description and 2D-grounding metrics, and obtains the best mAP and F1 for 3D grounding. Its 3D precision and closed-set planning accuracy remain competitive rather than being the best. The most notable gain over the conference version is on 2D grounding mAP, which improves by approximately 38\%. This result pattern is consistent with the main motivation of our design: distilling foundation visual-language features into explicit Gaussian primitives establishes finer correspondence between semantic concepts and spatial regions, while driving-aware selection and text-conditioned cross-attention preserve and aggregate task-relevant geometry into compact world tokens. Consequently, the improvement is most pronounced on tasks that require language to be accurately grounded in scene geometry, without compromising higher-level driving reasoning.

\noindent\textbf{NuInstruct.}
Tab.~\ref{tab:vqa-main-nuinstruct} evaluates cross-view risk perception, regression, classification, and captioning under the VGGDrive protocol. GaussianDWM++ achieves the best MAE, accuracy, mAP, and aggregate result, while its BLEU score remains competitive but is not the best. In particular, it reduces MAE by approximately 22\% relative to VGGDrive. NuInstruct requires the model to associate observations across views and reason about distances, motion states, and risk objects. The strong results on its geometric and prediction-oriented metrics therefore suggest that the continuous Gaussian representation preserves multi-view spatial relationships more effectively than image-level features alone. Meanwhile, the fact that the largest gains occur outside captioning indicates that the improvement primarily comes from better 3D-aware perception and reasoning rather than a stronger language prior.

\noindent\textbf{OmniDrive.}
As reported in Tab.~\ref{tab:vqa-main-omnidrive}, GaussianDWM++ ranks first on all language metrics and improves the best previous average result by approximately 14\%. Unlike NuInteract and NuInstruct, OmniDrive emphasizes holistic scene description and dialogue. The consistent improvement across these complementary settings indicates that the proposed representation is not limited to localization tasks. By anchoring foundation features to explicit 3D primitives and aligning the Gaussian world-token distribution with foundation image tokens, the model retains scene-level visual semantics while translating structured spatial evidence into language-compatible tokens. This leads to more complete and better-grounded descriptions of complex driving scenes.

The qualitative examples in Fig.~\ref{fig:experiments} further support this analysis by jointly visualizing the multi-view observations, 3D Gaussian scene representation, and language-based outputs. The responses remain tied to visible entities and their spatial relationships, illustrating how the shared Gaussian representation connects geometric evidence with semantic reasoning.

\begin{figure*}[!t]
    \centering
    \includegraphics[width=\linewidth]{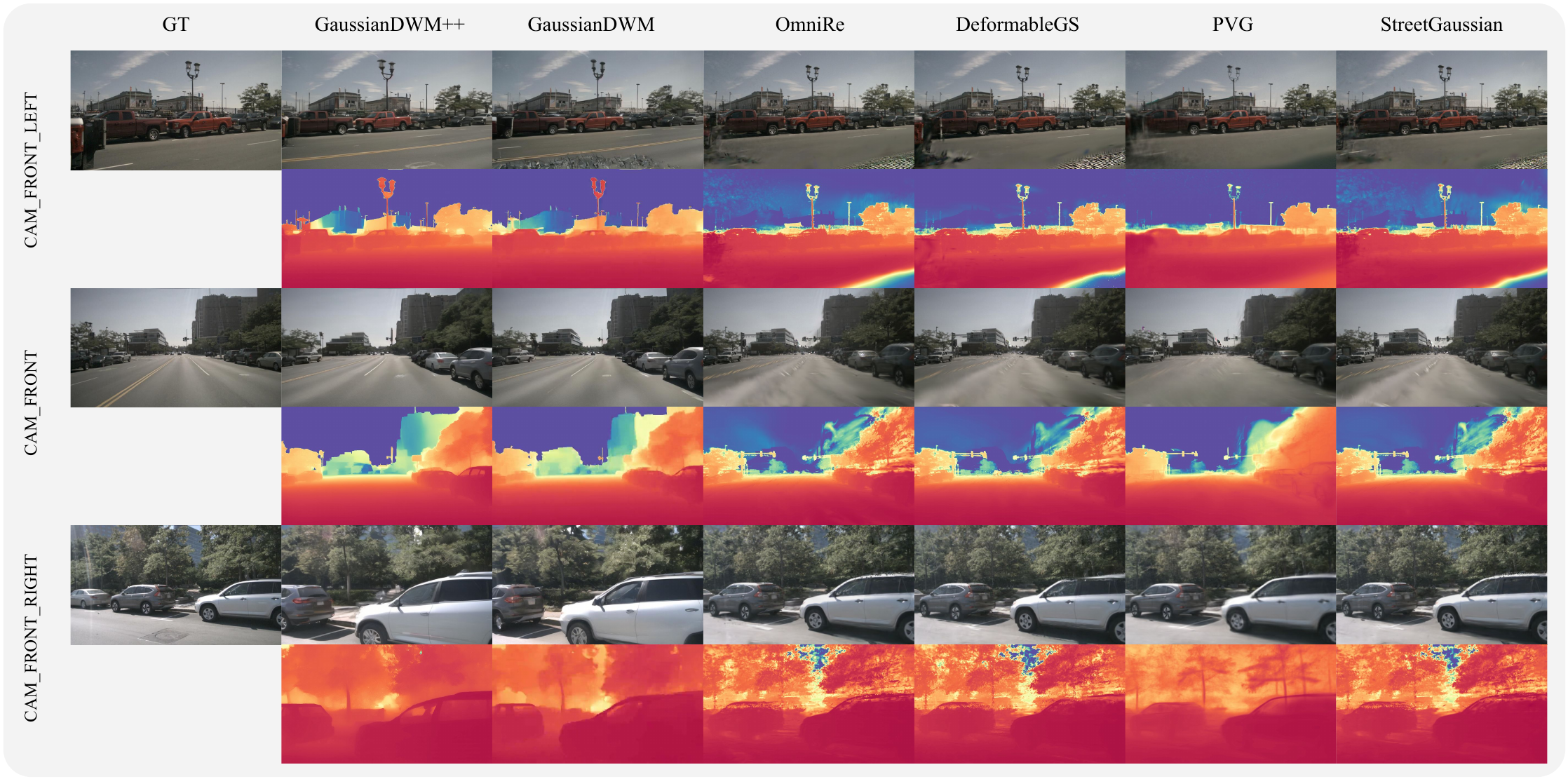}
    \vspace{-0.2cm}
    \caption{\textbf{Qualitative comparison of RGB-D NVS with a 4\,m camera shift.} Compared with GaussianDWM and reconstruction-based baselines~\cite{omnire,deformable3dgs,pvg,streetgaussian}, GaussianDWM++ better preserves broad, continuous scene regions such as road surfaces, with fewer large-area artifacts and more coherent depth structure under large viewpoint changes.
 }
    \vspace{-0.4cm}
    \label{fig:generation1}
\end{figure*}


\begin{table*}[!t]
\centering

\begin{threeparttable}
\resizebox{\textwidth}{!}{%
\begin{tabular}{llcccccccccccc}
\toprule
& & & \multicolumn{3}{c}{RD\&P $\uparrow$}
& \multicolumn{3}{c}{2D VG $\uparrow$}
& \multicolumn{3}{c}{3D VG $\uparrow$}
& \multicolumn{1}{c}{Plan $\uparrow$} & \\
\cmidrule(lr){4-6}\cmidrule(lr){7-9}\cmidrule(lr){10-12}\cmidrule(lr){13-13}
Model & Year & LLM & BLEU & ROUGE-L & CIDEr
& mAP & F1 & mIoU & Pr & mAP & F1 & Acc & Avg. $\uparrow$ \\
\midrule
LLaVA1.5~\cite{LLaVA} & 2024 & Vicuna-7B & 64.23 & 76.69 & 74.82 & 0.10 & 0.16 & 14.31 & 6.51 & 5.33 & 3.12 & 36.20 & 28.15 \\
MiniCPM-V 2~\cite{minicpm} & 2024 & MiniCPM-2B & 47.43 & 63.16 & 69.88 & 0.11 & 0.13 & 13.34 & 0.97 & 1.55 & 0.86 & 36.69 & 23.41 \\
MiniCPM-V 2.6~\cite{minicpm} & 2024 & Qwen2-7B & 47.92 & 69.11 & 70.20 & 0.36 & 0.49 & 18.74 & 1.97 & 1.61 & 0.93 & 36.42 & 24.78 \\
InternVL1.5-2B~\cite{internvl} & 2024 & InternLM2-7B & 67.14 & 81.10 & 79.83 & 14.74 & 17.64 & 55.43 & 28.05 & 21.73 & 12.92 & 53.96 & 43.25 \\
InternVL1.5-4B~\cite{internvl} & 2024 & Phi3-4B & 66.63 & 80.64 & 79.24 & 14.27 & 17.60 & 53.52 & 25.14 & 19.46 & 11.63 & 40.25 & 40.84 \\
Qwen2-VL-2B~\cite{qwen2} & 2024 & Qwen2-2B & 67.92 & 80.24 & 78.51 & 17.11 & 20.87 & 57.24 & 12.82 & 10.20 & 6.12 & 45.59 & 39.66 \\
Qwen2-VL-7B~\cite{qwen2} & 2024 & Qwen2-7B & 66.65 & 78.57 & 77.97 & 16.06 & 20.04 & 55.51 & 20.64 & 16.26 & 9.82 & 49.33 & 41.09 \\
InternVL2-1B & 2024 & Qwen2-0.5B & 66.89 & 81.00 & 79.59 & 16.70 & 20.21 & 55.94 & 23.36 & 18.35 & 10.94 & 44.08 & 41.71 \\
InternVL2-2B & 2024 & InternLM2-2B & 66.77 & 80.87 & 79.62 & 16.12 & 19.49 & 55.29 & 27.83 & 21.09 & 12.58 & 44.61 & 42.43 \\
InternVL2-4B & 2024 & Phi3-4B & 66.88 & 80.76 & 79.29 & 19.14 & 23.47 & 59.07 & 25.28 & 20.12 & 11.97 & 40.43 & 42.64 \\
InternVL2-8B & 2024 & InternLM2.5-7B & 67.32 & \ul{81.39} & \ul{80.01} & 20.61 & 25.24 & 61.90 & 31.47 & 24.67 & 14.70 & 46.93 & 45.42 \\
DriveMonkey~\cite{drivemonkey} & 2025 & InternLM2.5-7B & 67.50 & 81.15 & 79.79 & 19.47 & 24.02 & 59.36 & 51.90 & 34.53 & 20.86 & \textbf{82.64} & 52.12 \\
\midrule
BEVFormer~\cite{bevformer} & 2022 & -- & \multicolumn{6}{c}{--} & 44.50 & 23.69 & 1.58 & \multicolumn{2}{c}{--} \\
PETR~\cite{petr} & 2022 & -- & \multicolumn{6}{c}{--} & \textbf{55.80} & 31.34 & 20.58 & \multicolumn{2}{c}{--} \\
CAPE~\cite{cape} & 2023 & -- & \multicolumn{6}{c}{--} & \ul{55.02} & 32.94 & 21.33 & \multicolumn{2}{c}{--} \\
\midrule
GaussianDWM~\cite{Deng_2026_CVPR} & 2026 & Qwen3-8B
& \ul{68.78} & 81.06 & 78.72
& \ul{34.95} & \ul{40.49} & \ul{71.85}
& 50.66 & \ul{52.78} & \ul{32.05} & 80.95 & \ul{59.23} \\
\rowcolor{blue!8}
\textbf{GaussianDWM++} & 2026 & Qwen3-8B
& \textbf{69.81} & \textbf{81.75} & \textbf{80.42}
& \textbf{48.22} & \textbf{41.92} & \textbf{77.83}
& 53.32 & \textbf{57.19} & \textbf{35.13} & \ul{81.52} & \textbf{62.71} \\
\bottomrule
\end{tabular}}
\end{threeparttable}

\caption{\textbf{Comparison on NuInteract}~\cite{drivemonkey}. Results cover region description and perception (RD\&P), 2D/3D visual grounding (VG), and closed-set planning. Avg.\ is the arithmetic mean over the ten task metrics. All metrics are higher-is-better; best and second-best results are shown in \textbf{bold} and \ul{underlined}, respectively.}
\label{tab:vqa-main-nuinteract}
\end{table*}

\begin{table*}[!t]
\centering

\begin{threeparttable}
\resizebox{\textwidth}{!}{%
\begin{tabular}{cc*{9}{c}>{\columncolor{blue!8}}c}
\toprule
& & \multicolumn{4}{c}{Zero-shot Models}
& \multicolumn{6}{c}{Fine-tuned Models} \\
\cmidrule(lr){3-6}\cmidrule(lr){7-12}
Dataset & Metrics & GPT-4o~\cite{hurst2024gpt} & LLaVA-OV~\cite{li2024llava} & RoboTron~\cite{huang2025robotron} & Qwen2.5-VL~\cite{qwen25vl}
& Qwen2.5-VL~\cite{qwen25vl} & InMLLM~\cite{ding2024holistic} & VGGT-Dist~\cite{wang2026vggdrive} & VGGT-Add~\cite{wang2026vggdrive} & VGGDrive~\cite{wang2026vggdrive}
& \textbf{GaussianDWM++} \\
\midrule
\multirow{5}{*}{NuInstruct}
& MAE $\downarrow$ & 9.93 & 87.04 & 19.36 & 24.10 & 4.35 & 9.08 & 3.73 & 3.63 & \ul{3.08} & \textbf{2.41} \\
& Accuracy $\uparrow$ & 10.64 & 3.75 & 2.57 & 0.63 & 47.71 & 32.48 & 56.21 & 56.35 & \ul{56.37} & \textbf{58.02} \\
& mAP $\uparrow$ & 0 & 0 & 0 & 0 & 6.15 & 21.93 & 28.51 & 30.12 & \ul{37.49} & \textbf{42.59} \\
& BLEU $\uparrow$ & 7.08 & 8.55 & 8.06 & 5.56 & 75.75 & 35.20 & 79.23 & \ul{79.45} & \textbf{81.13} & 76.40 \\
& Average* $\uparrow$ & 1.95 & 0 & 0 & 0 & 31.32 & 20.13 & 40.06 & 40.57 & \ul{42.98} & \textbf{43.65} \\
\bottomrule
\end{tabular}}
\end{threeparttable}

\caption{\textbf{Comparison on NuInstruct}~\cite{ding2024holistic} under the VGGDrive evaluation
protocol. MAE is lower-is-better, while the remaining metrics are
higher-is-better. Average* is
$(\mathrm{Accuracy}+\mathrm{mAP}+\mathrm{BLEU}-\mathrm{MAE})/4$.
Best and second-best results are shown in \textbf{bold} and
\ul{underlined}, respectively.}
\label{tab:vqa-main-nuinstruct}
\end{table*}

\begin{table*}[!t]
\centering

\begin{threeparttable}
\resizebox{\textwidth}{!}{%
\begin{tabular}{cc*{9}{c}>{\columncolor{blue!8}}c}
\toprule
& & \multicolumn{3}{c}{Zero-shot Models}
& \multicolumn{7}{c}{Fine-tuned Models} \\
\cmidrule(lr){3-5}\cmidrule(lr){6-12}
Dataset & Metrics & GPT-4o~\cite{hurst2024gpt} & LLaVA-OV~\cite{li2024llava} & RoboTron~\cite{huang2025robotron} & OmniDrive~\cite{omnidrive}
& HERMES~\cite{hermes} & Qwen2.5-VL~\cite{qwen25vl} & VGGT-Dist~\cite{wang2026vggdrive} & VGGT-Add~\cite{wang2026vggdrive} & VGGDrive~\cite{wang2026vggdrive}
& \textbf{GaussianDWM++} \\
\midrule
\multirow{4}{*}{OmniDrive}
& BLEU $\uparrow$ & 10.91 & 16.14 & 20.30 & \ul{38.00} & -- & 37.29 & 37.28 & 37.23 & 37.58 & \textbf{42.11} \\
& CIDEr $\uparrow$ & 24.42 & 28.41 & 34.33 & 68.60 & 74.10 & 86.29 & 86.41 & 86.38 & \ul{86.57} & \textbf{100.00} \\
& ROUGE-L $\uparrow$ & 22.34 & 22.14 & 23.67 & 32.60 & 32.70 & 34.33 & 34.32 & 34.26 & \ul{34.40} & \textbf{38.13} \\
& Average $\uparrow$ & 19.22 & 22.23 & 26.10 & 46.40 & -- & 52.64 & 52.67 & 52.62 & \ul{52.85} & \textbf{60.08} \\
\bottomrule
\end{tabular}}
\end{threeparttable}

\caption{\textbf{Comparison on OmniDrive}~\cite{omnidrive} under the VGGDrive evaluation
protocol. Average is the arithmetic mean of BLEU, CIDEr, and ROUGE-L.
All metrics are higher-is-better. Best and second-best results are shown in
\textbf{bold} and \ul{underlined}, respectively.}
\label{tab:vqa-main-omnidrive}
\end{table*}

\subsection{Scene Generation}
We evaluate multi-modal scene generation on nuScenes~\cite{nuscenes}, including spatial novel-view generation under shifted camera poses and temporal prediction along the future ego trajectory. Following previous works~\cite{dist4d,magicdrive}, we interpolate the 2Hz keyframe annotations to 12Hz and adopt the DiST-S evaluation protocol. Quantitative spatial-generation results are reported in Tab.~\ref{tab:generation-main-spatial}, while qualitative spatial and temporal results are shown in Fig.~\ref{fig:experiments} and~\ref{fig:generation1}.

We compare with representative reconstruction- and generation-based approaches, including PVG~\cite{pvg}, EmerNeRF~\cite{emernerf}, StreetGaussian~\cite{streetgaussian}, OmniRe~\cite{omnire}, FreeVS~\cite{wang2025freevs}, and DiST-S~\cite{Guo2025ICCV}. As shown in Tab.~\ref{tab:generation-main-spatial}, GaussianDWM++ achieves the best FID and FVD under all three camera-shift settings. More importantly, its advantage becomes more pronounced as the viewpoint shift increases. Under the most challenging $\pm4$\,m setting, it reduces FID by approximately 20\% relative to the conference version. Large camera shifts substantially reduce view overlap and make image-only conditions increasingly ambiguous. In GaussianDWM++, the low-level RGB-D conditions projected from the explicit Gaussian scene provide persistent appearance and geometric constraints, while the high-level language condition supplies scene semantics when projected observations are sparse or incomplete. Their complementary roles explain why the method remains robust as the target view moves farther from the observed cameras.

The spatial and temporal examples in Fig.~\ref{fig:experiments} further show that the generated results preserve scene layout and maintain consistent appearance across target views and future frames. Since both conditions originate from the same foundation-feature Gaussian world representation, geometric evidence and semantic guidance remain mutually consistent rather than being fused from independently constructed scene representations. The qualitative RGB-D comparison in Fig.~\ref{fig:generation1} further supports this trend: GaussianDWM++ better preserves broad, continuous regions such as road surfaces under shifted viewpoints, reducing the large-area artifacts observed in GaussianDWM and reconstruction-based baselines. This behavior is consistent with the dual-condition design, where Gaussian-aware high-level world knowledge provides semantic context when projected RGB-D evidence becomes sparse or incomplete, complementing the low-level appearance and geometric constraints.

\begin{figure}
    \centering
    \includegraphics[width=\columnwidth]{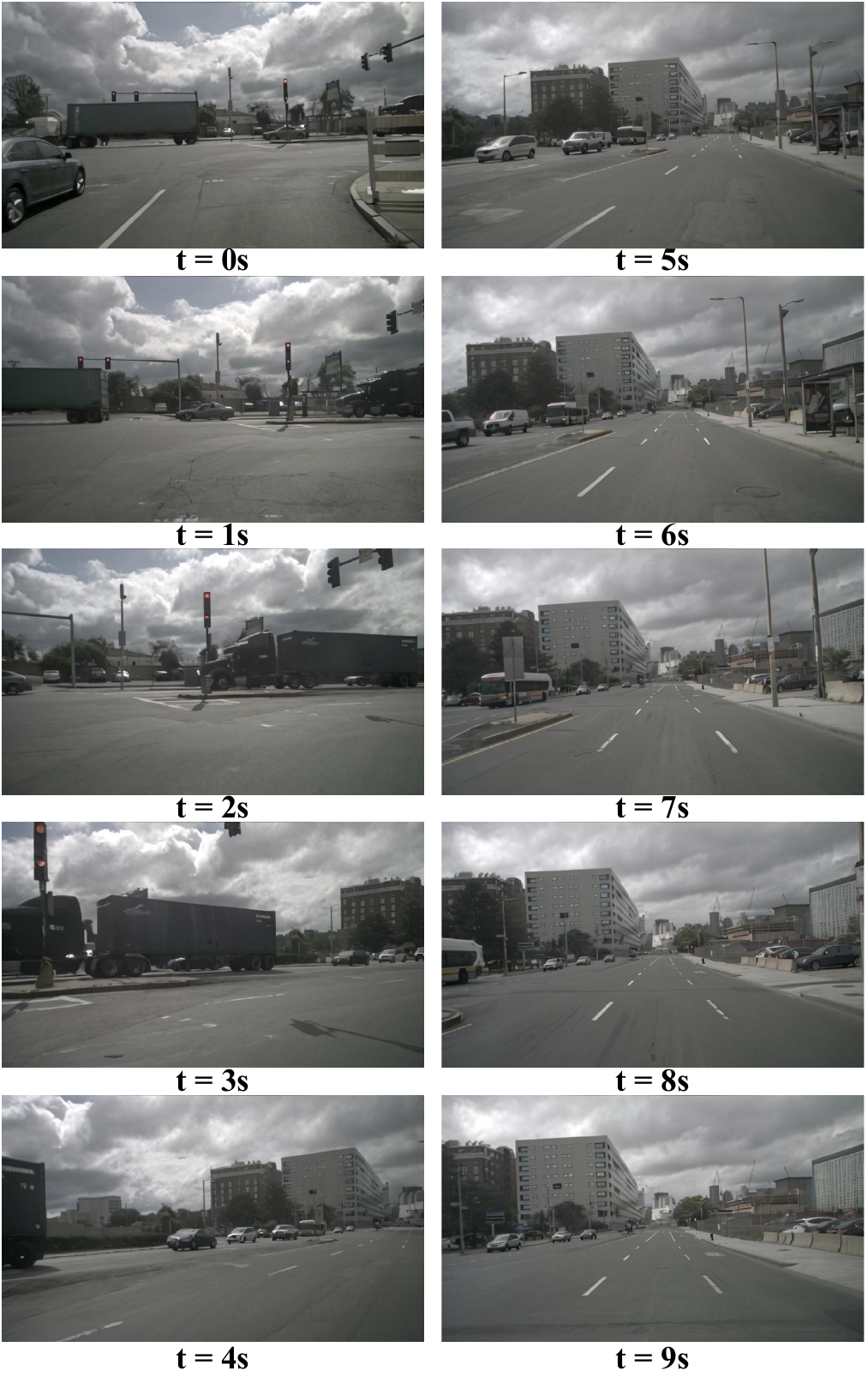}
\captionof{figure}{
Long-horizon generation from 0--9\,s.
}
\label{fig:generation-long-horizon}
\end{figure}

\begin{table}[htbp]
\centering
\begin{threeparttable}

\resizebox{\columnwidth}{!}{%
\begin{tabular}{lrrrrrr}
\toprule
& \multicolumn{2}{c}{$\pm1$\,m}
& \multicolumn{2}{c}{$\pm2$\,m}
& \multicolumn{2}{c}{$\pm4$\,m} \\
\cmidrule(lr){2-3}\cmidrule(lr){4-5}\cmidrule(lr){6-7}
Method & FID $\downarrow$ & FVD $\downarrow$
& FID $\downarrow$ & FVD $\downarrow$
& FID $\downarrow$ & FVD $\downarrow$ \\
\midrule
PVG~\cite{pvg} & 48.15 & 246.74 & 60.44 & 356.23 & 84.50 & 501.16 \\
EmerNeRF~\cite{emernerf} & 37.57 & 171.47 & 52.03 & 294.55 & 76.11 & 497.85 \\
StreetGaussian~\cite{streetgaussian} & 32.12 & 153.45 & 43.24 & 256.91 & 67.44 & 429.98 \\
OmniRe~\cite{omnire} & 31.48 & 152.01 & 43.31 & 254.52 & 67.36 & 428.20 \\
FreeVS~\cite{wang2025freevs} & 51.26 & 431.99 & 62.04 & 497.37 & 77.14 & 556.14 \\
DiST-S~\cite{Guo2025ICCV} & 10.12 & 45.14 & 12.97 & 68.80
& \ul{17.57} & \ul{105.29} \\
\midrule
GaussianDWM~\cite{Deng_2026_CVPR}
& \ul{8.36} & \ul{44.50}
& \ul{11.27} & \ul{68.17}
& 18.81 & 116.40 \\
\rowcolor{blue!8}
\textbf{GaussianDWM++}
& \textbf{8.19} & \textbf{42.65}
& \textbf{11.05} & \textbf{63.86}
& \textbf{15.02} & \textbf{101.20} \\
\bottomrule
\end{tabular}}
\end{threeparttable}

\caption{\textbf{Spatial novel-view generation on nuScenes.} FID and FVD
under $\pm1$\,m, $\pm2$\,m, and $\pm4$\,m camera shifts using the six-camera
DiST-S protocol; lower is better. Best and second-best results are shown in \textbf{bold} and
\ul{underlined}, respectively.}
\label{tab:generation-main-spatial}
\vspace{-3.5mm}
\end{table}

\begin{table}[!t]
\centering
\resizebox{\columnwidth}{!}{%
\begin{tabular}{lcccc}
\toprule
\textbf{Method}
& \textbf{6-View}
& \textbf{Video}
& \textbf{FID $\downarrow$}
& \textbf{FVD $\downarrow$} \\
\midrule

BEVGen
& \textcolor{ForestGreen}{\cmark}
& \textcolor{BrickRed}{\xmark}
& 25.54 & -- \\

BEVControl
& \textcolor{ForestGreen}{\cmark}
& \textcolor{BrickRed}{\xmark}
& 24.85 & -- \\

DriveGAN~\cite{drivegan}
& \textcolor{BrickRed}{\xmark}
& \textcolor{ForestGreen}{\cmark}
& 73.40 & 502.30 \\

DriveDreamer
& \textcolor{BrickRed}{\xmark}
& \textcolor{ForestGreen}{\cmark}
& 52.60 & 452.00 \\

Vista~\cite{vista}
& \textcolor{BrickRed}{\xmark}
& \textcolor{ForestGreen}{\cmark}
& 6.90 & 89.40 \\

\midrule

WoVoGen
& \textcolor{ForestGreen}{\cmark}
& \textcolor{ForestGreen}{\cmark}
& 27.60 & 417.70 \\

Panacea
& \textcolor{ForestGreen}{\cmark}
& \textcolor{ForestGreen}{\cmark}
& 16.96 & 139.00 \\

MagicDrive~\cite{magicdrive}
& \textcolor{ForestGreen}{\cmark}
& \textcolor{ForestGreen}{\cmark}
& 16.20 & 217.94 \\

DriveDreamer-2~\cite{drivedreamer}
& \textcolor{ForestGreen}{\cmark}
& \textcolor{ForestGreen}{\cmark}
& 25.00 & 105.10 \\

MagicDriveDiT
& \textcolor{ForestGreen}{\cmark}
& \textcolor{ForestGreen}{\cmark}
& 20.91 & 94.84 \\

Drive-WM~\cite{drivingwm}
& \textcolor{ForestGreen}{\cmark}
& \textcolor{ForestGreen}{\cmark}
& 15.80 & 122.70 \\

UniScene~\cite{uniscene}
& \textcolor{ForestGreen}{\cmark}
& \textcolor{ForestGreen}{\cmark}
& \textbf{6.12} & 70.52 \\

DiST-T~\cite{Guo2025ICCV}
& \textcolor{ForestGreen}{\cmark}
& \textcolor{ForestGreen}{\cmark}
& 6.83 & 22.67 \\


\midrule

\rowcolor{blue!8}
\textbf{GaussianDWM++}~\cite{Deng_2026_CVPR}
& \textcolor{ForestGreen}{\cmark}
& \textcolor{ForestGreen}{\cmark}
& \textbf{4.55} & \textbf{11.79} \\

\bottomrule
\end{tabular}%
}%
\caption{
\textbf{Temporal video generation on the nuScenes validation set.}
FID and FVD are lower-is-better. Best and second-best results are
shown in \textbf{bold} and \ul{underlined}, respectively.
}
\label{tab:generation-main-temporal}
\end{table}


The qualitative visualization in Fig.~\ref{fig:generation-long-horizon} and quantitative evaluation in Tab.~\ref{tab:generation-main-temporal} further assess the long-horizon generation capability of GaussianDWM++, demonstrating stable and temporally consistent world evolution over extended prediction horizons owing to the complementary dual-condition design and the incorporation of 3D world knowledge.

Overall, the quantitative results and the qualitative results establish state-of-the-art spatial generation performance and coherent future prediction. Together, they validate the dual-condition design and the use of a shared foundation-feature Gaussian representation to connect scene understanding with generation.

\subsection{Ablation Study}
We conduct ablation studies to examine how the Gaussian scene representation, token selection and alignment strategies, and dual-condition generation design contribute to the final performance. Rather than treating these components as interchangeable sources of improvement, we analyze the capability that each component primarily affects.


\begin{table*}[!t]
\centering

\begin{threeparttable}
\resizebox{\textwidth}{!}{%
\begin{tabular}{cccccccccccccc}
\toprule
\multirow{2}{*}{Fine-tuning} & \multirow{2}{*}{Gaussian}
& \multirow{2}{*}{Sampling}
& \multicolumn{3}{c}{RD\&P $\uparrow$}
& \multicolumn{3}{c}{2D VG $\uparrow$}
& \multicolumn{3}{c}{3D VG $\uparrow$}
& \multicolumn{1}{c}{Plan $\uparrow$} & \multirow{2}{*}{Avg. $\uparrow$} \\
\cmidrule(lr){4-6}\cmidrule(lr){7-9}\cmidrule(lr){10-12}\cmidrule(lr){13-13}
& & & BLEU & ROUGE-L & CIDEr & mAP & F1 & mIoU
& Pr & mAP & F1 & Acc & \\
\midrule
Zero-shot & No & --
& 2.91 & 12.68 & 0.59 & 0.00 & 0.00 & 12.24
& 48.75 & 47.59 & 29.12 & 0.00 & 15.39 \\
Fine-tuned & No & --
& 65.09 & 78.35 & 76.56 & 30.01 & 35.45 & 69.01
& 50.36 & 51.88 & 31.43 & 45.09 & 53.32 \\
Fine-tuned & Yes & Random
& \ul{66.19} & \ul{79.00} & \ul{76.97}
& \ul{33.94} & \ul{39.37} & \ul{71.40}
& \ul{50.94} & \ul{52.85} & \ul{32.03} & \ul{49.43} & 55.21 \\
Fine-tuned & Yes & Top-$K$ + Uniform
& \textbf{68.78} & \textbf{81.06} & \textbf{78.82}
& 33.89 & 39.31 & 71.37
& \textbf{51.16} & \textbf{52.87} & \textbf{32.05}
& \textbf{80.95} & \ul{58.93} \\
Fine-tuned & Yes & Top-$K$ + Uniform + Similarity
& \textbf{68.78} & \textbf{81.06} & \textbf{78.82}
& \textbf{34.95} & \textbf{40.49} & \textbf{71.85}
& 50.66 & 52.78 & \textbf{32.05} & \textbf{80.95} & \textbf{59.23} \\
\bottomrule
\end{tabular}}
\end{threeparttable}

\caption{\textbf{Ablation of the Gaussian scene representation and token
sampling on NuInteract.} The settings and results follow the conference-stage
evaluation protocol. Best and second-best results are shown in \textbf{bold} and
\ul{underlined}, respectively.}
\label{tab:qa-ablation-conference}
\end{table*}

\begin{table*}[!t]
\centering

\begin{threeparttable}
\resizebox{\textwidth}{!}{%
\begin{tabular}{cccc*{11}{c}}
\toprule
\multirow{2}{*}{\shortstack{Foundation-Feature\\Distillation}}
& \multirow{2}{*}{\shortstack{Coarse\\Selection}}
& \multirow{2}{*}{\shortstack{Fine\\Adapter}}
& \multirow{2}{*}{\shortstack{Distribution\\Alignment}}
& \multicolumn{3}{c}{RD\&P $\uparrow$}
& \multicolumn{3}{c}{2D VG $\uparrow$}
& \multicolumn{3}{c}{3D VG $\uparrow$}
& Plan $\uparrow$ & \multirow{2}{*}{Avg. $\uparrow$} \\
\cmidrule(lr){5-7}\cmidrule(lr){8-10}
\cmidrule(lr){11-13}\cmidrule(lr){14-14}
& & &
& BLEU & ROUGE-L & CIDEr
& mAP & F1 & mIoU
& Pr & mAP & F1
& Acc & \\
\midrule
\xmark
& Voxel Top-$K$
& Similarity
& \xmark
& 68.78 & 81.06 & 78.72
& 34.95 & 40.49 & 71.85
& 50.66 & 52.78 & 32.05
& \ul{80.95} & 59.23 \\
\xmark
& Voxel Top-$K$
& Similarity
& \cmark
& \ul{69.97} & 81.47 & 80.05
& \ul{48.59} & \ul{41.95} & 77.56
& \ul{53.39} & 57.18 & \ul{35.22}
& 67.02 & 61.24 \\
\xmark
& Voxel Top-$K$
& \shortstack{Query-Aware\\Cross-Attn}
& \cmark
& \textbf{70.19} & \ul{81.77} & \ul{80.11}
& 48.10 & 41.83 & 77.67
& \textbf{53.59} & \textbf{57.67} & \textbf{35.51}
& 67.21 & 61.37 \\
\xmark
& Driving-Aware
& \shortstack{Query-Aware\\Cross-Attn}
& \cmark
& 69.76 & \textbf{81.88} & 79.98
& \textbf{49.07} & \textbf{42.03} & \ul{77.76}
& 53.30 & 57.10 & 35.18
& 68.06 & \ul{61.41} \\
\cmark
& Driving-Aware
& \shortstack{Query-Aware\\Cross-Attn}
& \cmark
& 69.81 & 81.75 & \textbf{80.42}
& 48.22 & 41.92 & \textbf{77.83}
& 53.32 & \ul{57.19} & 35.13
& \textbf{81.52} & \textbf{62.71} \\
\bottomrule
\end{tabular}}
\end{threeparttable}

\caption{\textbf{Ablation of the foundation-feature Gaussian tokenizer and
geometry-aware Gaussian adapter on NuInteract.} Rows progressively introduce
KL-based Gaussian--image distribution alignment, text-conditioned
cross-attention, and driving-aware coarse selection. The last row additionally
enables foundation-feature distillation.
Avg.\ is the arithmetic mean over the ten task metrics; higher is better.}
\label{tab:qa-ablation-journal}
\end{table*}


\noindent\textbf{Gaussian Representation and Hybrid Sampling.}
Tab.~\ref{tab:qa-ablation-conference} first examines the conference-stage Gaussian representation and sampling strategy. Fine-tuning is essential for adapting the language model to driving tasks, while introducing Gaussian tokens consistently improves the fine-tuned image-only baseline across description, grounding, and planning. The comparison among sampling strategies further shows that the benefit does not simply come from providing more 3D primitives. Replacing random sampling with Top-$K$ plus uniform sampling produces the most pronounced gain in planning accuracy, with an improvement of approximately 64\%. This indicates that a mixture of salient primitives and broad spatial coverage is important for retaining the global context required by driving decisions. Adding text--Gaussian similarity selection yields the best overall result and the strongest 2D-grounding performance, whereas Top-$K$ plus uniform sampling remains slightly better on several 3D-grounding metrics. Thus, query relevance mainly refines object-level localization, while geometry-preserving coverage remains important for metric 3D reasoning.

\noindent\textbf{Foundation-Feature Tokenizer and Geometry-Aware Adapter.}
Tab.~\ref{tab:qa-ablation-journal} further separates the journal-version components. Introducing Gaussian--image distribution alignment produces the largest change in 2D grounding, improving mAP by approximately 39\%, and also benefits most description and 3D-grounding metrics. This supports the role of the KL objective in transferring foundation image-token semantics to Gaussian world tokens. However, distribution alignment alone does not preserve the best planning result, showing that image-guided compatibility is necessary but insufficient for high-level driving reasoning. Text-conditioned cross-attention achieves the strongest 3D-grounding results, while driving-aware coarse selection performs best on the principal 2D-grounding metrics. These complementary trends agree with the adapter design: coarse selection maintains driving-scene coverage, whereas text-modulated learned queries aggregate task-relevant primitives into a fixed number of world tokens. Finally, enabling foundation-feature distillation improves planning accuracy by approximately 20\% over the preceding configuration and produces the best aggregate result. Although the complete model does not rank first on every individual metric, it provides the best balance among semantic description, spatial grounding, and planning-oriented reasoning.

\noindent\textbf{Dual-Condition Generation.}
Tab.~\ref{tab:world} analyzes the roles of low-level geometric observations and high-level world knowledge. High-level conditioning alone fails to generate valid results because semantic guidance cannot determine the target-view geometry and appearance without a spatial anchor. Low-level RGB-D conditioning is sufficient to make generation feasible, but combining it with the high-level condition further reduces FID by approximately 17\%. This confirms that the two conditions play complementary rather than redundant roles: projected RGB-D evidence constrains scene layout and local appearance, while Gaussian-aware language features supply semantic context for uncertain or incomplete regions. The ablation therefore explains why the full model improves perceptual quality without sacrificing geometric consistency.

\begin{table}[!t]
\setlength{\tabcolsep}{1.1mm}
\centering
\begin{tabular}{ll|ll}
\hline
Low-Level & High-level & FID $\downarrow$ & FVD $\downarrow$ \\ \hline
\ding{55}      & $\checkmark$       & -       & -       \\
$\checkmark$     & \ding{55}        & 10.12      & 45.14      \\ \hline
$\checkmark$     & $\checkmark$       & \textbf{8.36}       & \textbf{44.5}       \\ \hline
\end{tabular}
\caption{\textbf{Ablation study of the dual-condition generation mechanism under a $\pm1$\,m camera shift.} FID and FVD are lower-is-better. ``--'' denotes failure under the corresponding setting.}
\label{tab:world}
\end{table}

\section{Conclusion}
In this paper, we present GaussianDWM++, a foundation-feature 4D Gaussian driving world model for unified scene understanding, language-grounded reasoning, controllable editing, and multi-modal generation. By directly distilling Qwen features into 3D Gaussian primitives, our framework builds an explicit open-vocabulary Gaussian scene representation and reduces the dependence on external language Gaussian construction pipelines. To efficiently bridge dense 3D Gaussians with large language models and generative models, we further introduce a geometry-aware Gaussian adapter with importance-aware sampling, text-conditioned Perceiver-style cross-attention, and KL-based Gaussian--image distribution alignment. Based on the aligned Gaussian world representation, our model supports controllable 4D scene editing, including weather-conditioned generation and dynamic vehicle manipulation. We also construct NuScenes-GSQA with 1000 foundation-feature Gaussian driving scenes and Gaussian-aware QA annotations, providing a useful benchmark for future 3D Gaussian-based driving world models. Extensive experiments demonstrate that the proposed framework effectively improves scene understanding, visual grounding, planning-oriented reasoning, and controllable multi-modal generation, showing the potential of 3D Gaussian representations as a unified foundation for future driving world models.







\bibliographystyle{IEEEtran}
\bibliography{egbib}

\end{document}